\documentclass[runningheads]{llncs}

\usepackage[noend]{algpseudocode}
\usepackage{algorithm}
\usepackage{graphicx}
\usepackage{hyperref}
\usepackage{colortbl}
\usepackage{threeparttable}
\usepackage{mathtools}

\algdef{SE}[DOWHILE]{Do}{doWhile}{\algorithmicdo}[1]{\algorithmicwhile\ #1}%
\usepackage{mathtools}
\usepackage{graphicx} % Allows including images
\usepackage{booktabs}
\usepackage{multirow}
\usepackage{multicol}
\usepackage{caption}
\usepackage{subcaption}

\usepackage[utf8]{inputenc} % allow utf-8 input
\usepackage[T1]{fontenc}    % use 8-bit T1 fonts
\usepackage{url}  % simple URL typesetting
\usepackage{makecell}
\usepackage{booktabs}       % professional-quality tables
\usepackage{amsfonts}       % blackboard math symbols
\usepackage{nicefrac}       % compact symbols for 1/2, etc.
\usepackage{microtype}      % microtypography
\usepackage{lipsum}
\usepackage{amsmath}

\newcommand\numberthis{\addtocounter{equation}{1}\tag{\theequation}}
\usepackage{pgfplotstable}
\usepackage{soul}
\usepackage{amssymb}% http://ctan.org/pkg/amssymb
\usepackage{pifont}% http://ctan.org/pkg/pifont
\newcommand{\cmark}{\ding{52}}%
\newcommand{\xmark}{\ding{53}}%

\usepackage{cleveref}% http://ctan.org/pkg/cleveref
\usepackage{lipsum}% http://ctan.org/pkg/lipsum
\begin{document}
\title{Overlapping Community Detection using Dynamic Dilated Aggregation in Deep Residual GCN}

\author{Md Nurul Muttakin\inst{1}\orcidID{0000-0002-6114-6260} \and
Md. Iqbal Hossain\inst{2}\orcidID{0000-0001-6212-7638} \and
Md. Saidur Rahman\inst{1}\orcidID{0000-0003-0112-0242}}
\authorrunning{N. Muttakin et al.}
\institute{Bangladesh University of Engineering and Technology, Dhaka 1000, Bangladesh 
 \and
University of Arizona, USA}
\maketitle  
%% - \title{} in title case
%% - \Plaintitle{} without LaTeX markup (if any)
%% - \Shorttitle{} with LaTeX markup (if any), used as running title

%% - \Abstract{} almost as usual
\begin{abstract}
Overlapping community detection is a critical problem in unsupervised machine learning on network-structured data. Some research has considered applying graph convolutional networks (GCN) to tackle the problem. However, the considered GCN is shallow and can not capture communities with larger diameters. Moreover, it is still open to incorporate dynamic dilated aggregation in deep GCNs for irregular graphs. In this study, we design a deep dynamic residual GCN (DynaResGCN) based on our dynamic
dilated aggregation algorithm and a unified end-to-end encoder-decoder-based framework to detect overlapping communities. The deep DynaResGCN model is used as the encoder, whereas we incorporate the Bernoulli-Poisson (BP) model as the decoder.  Consequently, we apply our overlapping community detection framework in a research topics dataset without having ground truth, a set of networks from Facebook having reliable (hand-labeled) ground truth, and in a set of very large co-authorship networks having empirical (not hand-labeled) ground truth. Our experimentation on these datasets shows significantly superior performance over many state-of-the-art methods for detecting overlapping
communities in networks.

\keywords{Overlapping Community Detection \and Deep GCN \and Unsupervised Learning}
\end{abstract}

%% - \Keywords{} with LaTeX markup, at least one required
%% - \Plainkeywords{} without LaTeX markup (if necessary)
%% - Should be comma-separated and in sentence case.

% \Plainkeywords{Overlapping Community Detection, Deep GCN, Unsupervised Learning}

%% - \Address{} of at least one author
%% - May contain multiple affiliations for each author
%%   (in extra lines, separated by \emph{and}\\).
%% - May contain multiple authors for the same affiliation
%%   (in the same first line, separated by comma).
% \Address{
%   Md Nurul Muttakin \\ 
%   Graph Drawing and Information Visualization Laboratory \\
%   Department of Computer Science and Engineering \\
%   Bangladesh University of Engineering and Technology \\
%   Dhaka-1000 \\
%   E-mail: \email{1018052056@grad.cse.buet.ac.bd}
% \\
% \\
%   Md Iqbal Hossain \\
%   Research, Innovation and Impact \\ 
%   University of Arizona, USA \\
%   E-mail: \email{hossain@arizona.edu}
%    \\
%   \\
%   Md Saidur Rahman\\
%   Graph Drawing and Information Visualization Laboratory\\
%   Department of Computer Science and Engineering\\
%   Bangladesh University of Engineering and Technology\\
%   Dhaka-1000\\
%   E-mail: \email{saidurrahman@cse.buet.ac.bd}
% }

% \pgfplotsset{compat=1.18}

\section{Introduction}
%\textcolor{red}{
Network %/graph Graph
structured data are ubiquitous. Almost every domain of science and engineering has to deal with networks. %structured data.
The nodes of real-world networks often form special groups where %edge/
link density is high, and the density of edges is low among these groups. This property of networks is called \textit{community structure} \cite{girvan2002community}. Human society is composed of communities that are in many cases %online/
virtual due to the wide use of online social platforms. Indeed, social communities are of great importance to social scientists and have been studied thoroughly for a long time \cite{coleman1964introduction-comm-math-sociology,wellman2008development-social-network-analysis,herskovits1955cultural-social-com,moody2003structural-social-com}. Communities in the global financial network play a key role in influencing the transitions of the global economy and financial system \cite{chan2018systemic-financial-com}. In protein-protein interaction networks, protein communities act as functional modules to perform specific tasks in cells \cite{rives2003modular-ppi-network,spirin2003protein-network-2,chen2006detecting-com-ppi}.
% \todo{Organize this part for more general example of graphs (not specific to only social networks)}
% Problem definition
% Problem motivation,
% Community detection in graphs is basically detection of group of nodes which have higher connectivity than the nodes outside of the group. Communities are not disjoint always. In case of 
%In a nutshell,
As a consequence, %}
community detection %is necessary to analyze
%requires analysis of social networks %and 
is an essential tool to analyze different kinds of networks  such as social networks, research topics networks, protein-protein interaction networks, ecological networks, financial networks, etc.,
and to understand %\textcolor{red}{/interpret} 
their structure.  %\cite{girvan2002community}.

The communities in most real-world networks, especially social networks, overlap \cite{yang2014structure}. The same participant participates in different communities at the same time.
Sometimes, a participant in a community might simultaneously be a different community member. A person in a social network is in his or her family's community while he/she is also in the community of his or her coworkers. In reality, a node may belong to an unlimited number of communities in a social network. This overlapping node is critical for various reasons, as it might be used positively and negatively. For instance, in a social network context, a node belonging to two disputing communities might be potentially used to resolve the dispute. On the contrary, this node may disclose sensitive information %of
about a community to other communities violating privacy ethics. 
Indeed, the overlap is an inherent characteristic of many real-world social networks \cite{kelley2012definingOverlap}. Moreover, disjoint community detection induces a partition in the network that eventually breaks the communities \cite{reid2013partitioning}. 
% Additionally, overlapping community detection generalizes disjoint community detection. For example, if an overlapping community detection method is applied on a network with highly disjoint communities, the method should return very few nodes having multiple community memberships.
Therefore, detecting overlapping communities in different networks is of utmost importance, %, especially in the case of social networks.
and thus, it has attracted significant attention in the research community \cite{xie2013overlapping-survey}.

% Literature,
% \todo{Make this part more coherent and smoother}

Many %of the 
works in the literature are related to non-overlapping community detection \cite{abbe2017community,von2007tutorial}. In the case of non-overlapping community detection, node embedding approaches perform well %to detect non-overlapping communities
\cite{tsitsulin2018verse,cavallari2017learning}. %\cite{chen2017supervised-line-gnn}. %Chen et al. \cite{chen2017supervised} performed supervised community detection with line graph neural networks. %which is only non-overlapping community detection.
However, most of these approaches are not well suited for overlapping community detection %\textcolor{red}{
due to the lack of scalable and reliable clustering methods %or algorithms
for high-dimensional vectors. A viable approach to resolve this issue is to generate the node embedding as community affiliation, which can be done effectively and efficiently using GCNs %}
\cite{shchur2019overlapping}. 

Among the classical approaches for overlapping community detection, label propagation algorithms are prominent \cite{mahabadi2021slpa-para}. For instance, SLPA is an overlapping community detection algorithm based on the idea of label propagation \cite{xie2011slpa}. DEMON \cite{coscia2014uncovering-demon}, a node-centric bottom-up overlapping community detection algorithm, also leverages ego network structure %and
with overlapping label propagation. Other algorithmic approaches include seed-set expansion \cite{asmi2021greedy-seed,gao2021overlapping-pagerank}, evolutionary algorithms \cite{ma2021local-evolutionary}, and heuristics-based methods \cite{ruan2013efficient,galbrun2014overlapping,gleich2012vertex,li2015uncovering}. %Heuristics-based approaches were considered by many works such as Ruan et al. \cite{ruan2013efficient}, Galbrun et al. \cite{galbrun2014overlapping}, Gleich and Seshadhri \cite{gleich2012vertex}, and Todeschini et al. \cite{li2015uncovering}. 
Some heuristics-based works considered game-theoretic concepts \cite{wang2021effective-game}.

Many studies consider some kinds of matrix factorization along with probabilistic inference \cite{yang2013overlapping-bigclam,zhou2015infinite-epm,todeschini2016exchangeable-snetoc,latouche2011overlapping,kuang2012symmetric,li2018community-cde,wang2011community-snmf}. One of the earliest works in this regime is BigClam \cite{yang2013overlapping-bigclam}.
BigClam was designed for overlapping community detection in large networks \cite{yang2013overlapping-bigclam}. BigClam uses gradient ascent to create an embedding, which is later used to determine nodes' community affiliation. MNMF \cite{wang2017community-mnmf} uses modularity-based regularization with non-negative matrix factorization. DANMF uses a telescopic non-negative matrix factorization approach based on a deep auto-encoder to learn membership distribution over nodes \cite{ye2018deep-danmf}.  Some works \cite{cao2018incorporating,yang2016modularity} performed factorization of modularity matrices using neural nets, %Some researchers  \cite{} performed overlapping community detection based on a non-negative matrix factorization approach in different settings. Many studies  developed overlapping community detection algorithms based on probabilistic inference, and  
while other works considered adversarial learning  \cite{jia2019communitygan,chen2021self-adv} or deep belief networks \cite{hu2017deep} to learn community affiliation. However, they did not use graphs in their neural network architecture, which is crucial to achieving superior performance \cite{shchur2019overlapping}.  Finally, NOCD \cite{shchur2019overlapping} used %special neural networks called 
a graph neural networks (GNN) based encoder-decoder model to detect overlapping communities. Unlike NMF-based methods, it optimizes the weights of a GNN to generate a better community affiliation matrix.

% Our Contribution,

Graph neural networks (GNN) are a class of specialized neural networks %which
that can operate on graph-structured data \cite{hamilton2017inductive,kipf2016semi,xu2018representation}. %In general, 
% GNNs use  a special kind of convolution operation called graph convolution \cite{kipf2016semi}. In graph convolution, each node aggregates information from the neighboring nodes. Before aggregation, the information of neighboring nodes is transformed by a learnable kernel. One of the major goals of graph learning is to generate node embedding to perform downstream induction tasks, such as node classification, edge labeling, clustering, or community detection \cite{perozzi2014deepwalk, berg2017graph, grover2016node2vec, bojchevski2017deep-g2g}.  
The authors of NOCD \cite{shchur2019overlapping} first developed a graph neural network-based approach for overlapping community detection. In that approach, they used a %2 
two-layer shallow graph convolutional network as an encoder to generate community embedding and a decoder based on the %\textcolor{red}{
Bernoulli-Poisson %}
model \cite{yang2013overlapping-bigclam,zhou2015infinite-epm,todeschini2016exchangeable-snetoc} to reconstruct the original graph from the embedding. However, a shallow GCN %\textcolor{red}{ 
%which 
has essential limitations to capturing  %learn %/represent
communities with a larger community diameter \cite{ghoshal2017diameter} ($>2$) as it can aggregate information only up to two hops from a node.
% \emph{Community diameter} \cite{ghoshal2017diameter} refers to the maximum shortest path distance between any two nodes of a community-induced subgraph (Figure \ref{fig:comm-diameter}). %}
% \begin{figure}[h]
%     \centering
%     \includegraphics[width=0.50\textwidth]{images/com-diameter-math-all.pdf}
%     \caption{
%     Two overlapping communities with community diameter $>2$.
%     }
%     \label{fig:comm-diameter}
% \end{figure}
One solution to the issue is to consider a deep graph convolutional networks (DeepGCN) \cite{li2019deepgcns,chen2020simple-deepgcn,valsesia2020deepgcn-image,luan2019break-deepgcn-nips} %proposed by Li et al. \cite{li2019deepgcns}
instead of a shallow  GCN. However,  Li et al. \cite{li2019deepgcns} %first 
proposed %a deep graph convolutional network (
the DeepGCN in the case of point cloud learning, where the considered graph is regular while the real-world networks are mostly irregular graphs (having variable-sized neighborhoods). The DeepGCN proposed by \cite{valsesia2020deepgcn-image} is also specific to regular graphs where image pixels are considered nodes, while \cite{chen2020simple-deepgcn} completely ignores dilated aggregation. Moreover, \cite{luan2019break-deepgcn-nips} proposes a different approach that requires QR factorization with backpropagation, making it intractable for practical implementation. Thus, it is %required 
essential to design a deep GCN that can handle the irregular graphs %having variable sized neighborhoods 
to detect overlapping communities %having 
with a larger diameter.
%This work motivated us to consider deep graph neural networks in the case of overlapping community detection, since the previous work \cite{shchur2019overlapping} NOCD considered %only
% a shallow GCN %\textcolor{red}{ 
% which has essential limitation to capture  %learn %/represent
% communities with larger diameter ($>2$) as it can aggregate information only upto two hops from a node. \emph{Community diameter} refers to the maximum shortest path distance between any two nodes of a community induced subgraph (Figure \ref{fig:comm-diameter}). %}

Training %a 
deeper GCNs is challenging because of over-smoothing and vanishing gradient problems \cite{li2019deepgcns,chen2020simple-deepgcn,valsesia2020deepgcn-image,luan2019break-deepgcn-nips}. Previous works \cite{li2019deepgcns,valsesia2020deepgcn-image,chen2020simple-deepgcn} overcame this challenge %in %the case of 
%point cloud learning by borrowing
with the idea of residual connection and dilated aggregation borrowed from %a classical CNN-based architecture called 
ResNet \cite{he2016deepResNet}. %\textcolor{red}{
\emph{Residual connection}
%/skip connection}}
means having direct connectivity from a layer's input to that layer's output, bypassing the non-linearity in the
activation function. In %the case of 
GCNs, each node aggregates information from the neighboring nodes. It is called \emph{dilated aggregation} if this aggregation mechanism aggregates information from some of the distant neighbors while skipping some of the nearest
neighbors. 
In %the case of 
dilated aggregation, %it is possible to
we may incorporate some %sort of 
randomization while considering neighbors, which introduces the concept of %\textcolor{red}{\textit
\emph{edge dynamicity}. Dilated aggregation with edge dynamicity is called \emph{dynamic dilated aggregation}.
%In the case of point cloud learning using deep GCNs, a graph is generated from the point cloud using k-nearest neighbor algorithm \cite{li2019deepgcns}. In this generated graph, each node has the same number of neighbors. However, in the general \textcolor{red}{irregular} graphs, the number of neighbors for each node varies widely. %Moreover, in case of weighted graphs, it is crucial to incorporate the information coming from weights of edges. 
It is still challenging to introduce dynamic dilated aggregation in %the case of %\textcolor{red}{
irregular %} 
graphs %(having variable length neighborhood) %and incorporate information from weights of edges in order 
%in order 
to train deep GCNs. Indeed, none of the existing works on deep GCN considered dilated aggregation in \textit{irregular} graphs.
%\textcolor{red}{ In Table \ref{tab:compare_with_others}, we provide  a conceptual comparison of our method with the relevant literature.}

In this study, we % have designed 
design %different %novel 
%methods of 
dynamic dilated aggregation mechanisms in %the case of 
irregular graphs. %considering both weighted and unweighted edges.
We incorporate our dynamic dilated aggregation schemes along with residual connection %to 
into the GCN to obtain a deep GCN that %is able to
can learn from any graphs. %(considering variable length neighborhood). 
Eventually, we apply our designed deep GCN to detect overlapping communities in different kinds of networks with different %scales/
sizes. %successfully. 
%In the case of
For overlapping community detection, we adopt an encoder-decoder-based approach similar to NOCD. %\cite{shchur2019overlapping}. 
Our encoder is a deep GCN model called DynaResGCN, which generates the community embedding, and the decoder attempts to reconstruct the original graph. 

Our objective is summarized as follows:
\begin{itemize}
    \item Detecting overlapping communities with higher community diameters effectively
    \item Designing a DeepGCN model 
\end{itemize}
 In Table \ref{tab:compare_with_others}, we provide a conceptual comparison of our method with the relevant literature.
%Then we applied our model with deep GCN to detect overlapping communities in a large research topic network. 

%Then we applied our model with deep GCN to detect overlapping communities in a large research topic network. 

%To evaluate our models
%In the case of 
\begin{table}[h]
\setlength\tabcolsep{3.5pt} % default value: 6pt
%   \begin{threeparttable}[b]
    \centering
%     {\footnotesize
% 	\renewcommand\arraystretch{1.8}
	\caption{
	Conceptual comparison of our method to the relevant literature.
    }
  \label{tab:compare_with_others}
	\centering 
	\resizebox{\columnwidth}{!}{
	\begin{tabular}{cccccccc}\toprule
		\makecell{Algorithm/\\Model} & \makecell{Overlapping \\ community \\ detection } & \makecell{Large \\ community\\ diameter} & \makecell{Residual \\ connection}&\makecell{Edge \\ dynamicity}  &  \makecell{Dil.Agg. \\ in irregular \\ graphs$^1$} & \makecell{Deep \\GCN\\ model} & Reference \\
		\midrule
		DynaResGCN+BP & \cmark & \cmark  & \cmark & \cmark & \cmark & \cmark & \textbf{Ours} \\
		NOCD & \cmark  & \xmark   & \xmark & \xmark & \xmark & \xmark & {\cite{shchur2019overlapping}} \\
		BigClam & \cmark  & \cmark & \xmark & \xmark & \xmark & \xmark & {\cite{yang2013overlapping-bigclam}} \\
		SLPA & \cmark  & \cmark & \xmark & \xmark & \xmark & \xmark & {\cite{xie2011slpa}} \\
		DEMON & \cmark  & \cmark & \xmark & \xmark & \xmark & \xmark & {\cite{coscia2014uncovering-demon}} \\
		DANMF & \cmark  & \cmark & \xmark & \xmark & \xmark & \xmark & {\cite{ye2018deep-danmf}} \\
		CDE & \cmark  & \cmark & \xmark & \xmark & \xmark & \xmark & {\cite{li2018community-cde}} \\
		EPM & \cmark  & \cmark & \xmark & \xmark & \xmark & \xmark & {\cite{zhou2015infinite-epm}} \\
		CESNA & \cmark  & \cmark & \xmark & \xmark & \xmark & \xmark & {\cite{yang2013community-cesna}} \\
      SNMF & \cmark  & \cmark & \xmark & \xmark & \xmark & \xmark & {\cite{wang2011community-snmf}} \\
          SNetOC & \cmark  & \cmark & \xmark & \xmark & \xmark & \xmark & {\cite{todeschini2016exchangeable-snetoc}} \\
  DeepWalk with K-means & \cmark  & \cmark & \xmark & \xmark & \xmark & \cmark & {\cite{perozzi2014deepwalk}} \\

  Graph2Gauss with K-means& \cmark  & \cmark & \xmark & \xmark & \xmark & \xmark & {\cite{bojchevski2017deep-g2g}} \\
  DeepGCN & \xmark & N/A & \cmark & \cmark & \xmark & \cmark & {\cite{li2019deepgcns}} \\
   GCDN & \xmark & N/A & \cmark & \cmark & \xmark & \cmark & {\cite{valsesia2020deepgcn-image}} \\
   GCNII & \xmark & N/A & \cmark & \xmark & \xmark & \cmark & {\cite{chen2020simple-deepgcn}} \\
    Krylov GCN & \xmark & N/A & N/A & N/A & N/A & \cmark & {\cite{luan2019break-deepgcn-nips}} \\
		
		\bottomrule
    \end{tabular}
    }
    \begin{tablenotes}
    {
    % \scriptsize
    \footnotesize
        
        \item (1) Dilated aggregation in the graphs having variable-sized neighborhoods. 
        %}
      }
    \end{tablenotes}
%   \end{threeparttable}
\end{table}
For evaluation, we consider a large research topics network \cite{burd2018gram}, a set of Facebook datasets \cite{mcauley2014discovering-facebook-dataset} having reliable ground truth information, and a set of very large co-authorship networks (nodes $> 10K$) \cite{shchur2019overlapping} as an additional benchmark. %for overlapping community detection in very large graphs 
 In the case of the topics network, ground truth information related to overlapping communities does not exist. Therefore, we have to rely on unsupervised metrics called fitness functions or quality measures %,
such as conductance, clustering coefficient, density, and coverage  \cite{malhotra-lpa2021modified,yang2013overlapping-bigclam} %, 
to determine the quality of the detected overlapping communities. In addition to the evaluation metrics, we generate heatmaps or colormaps to illustrate community overlap. We compare our results in topics network with existing overlapping community detection approaches, such as %like 
NOCD, %\cite{shchur2019overlapping}
 BigClam, %\cite{yang2013overlapping-bigclam}
 SLPA, %\cite{xie2011slpa}
 DANMF, % \cite{ye2018deep-danmf},
 and DEMON. %\cite{coscia2014uncovering-demon}.
 Our methods show superior performance over most of the considered existing methods %on 
in terms of both quality metrics and visualization. In the case of the datasets having ground truth information, we compare with  exactly the same baselines as in NOCD %\cite{shchur2019overlapping}
considering both node features and without node features in terms of normalized mutual information (NMI) \cite{mcdaid2011normalized-nmi}. Our method, DynaResGCN, achieves a clear superior performance over all the baselines in almost every case in terms of NMI. %\textcolor{red}{ In essence,
In summary, our contributions are %summarised 
as follows:
\begin{itemize}
    
    \item %Developed
    Development of a deep %dynamic
    residual %graph convolutional network
    GCN (DynaResGCN) model based on %novel
    dynamic dilated aggregation %which can operate on graphs having variable-sized neighborhood (irregular graph)
    in \emph{irregular} graphs 
    \item Design of an %highly 
     effective and scalable overlapping community detection framework based on %developed deep 
    DynaResGCN and Bernoulli-Poisson model% which is 
    \item Detection of overlapping communities in real-world networks outperforming many state-of-the-art methods
    % \item Devised different approaches to incorporate information coming from edge-weights of weighted graphs in DynaResGCN
\end{itemize}

% Paper Organization,
% (Whole paper summary).
\par
The remainder of the paper is organized as follows. In Section 2,
we introduce preliminary terminologies and some of the prominent related works. We describe our methodology in Section 3. Section 4 deals with our experimentation. In section 5, we present our results and discuss our insights with overall findings. Finally, Section 6 %is a 
provides our conclusions. 
% In preliminaries we discuss basic concepts, ideas, definitions related to our paper. In the methodology section we elaborate detailed on our methodologies. We basically adopted two previous methods on graph neural networks: dilation filtering and neighborhood aggregation. Finally, we present our experimental methods, results and conclusions. 

\section{Preliminaries}
In this section, we %give 
introduce the necessary notations and terminologies that will be used throughout the paper and discuss prominent related works briefly. %We use the same notations as described %as  in \cite{shchur2019overlapping}. 
\subsection{Notations and Terminologies}
We consider an undirected %unweighted
graph as $G=(V, E)$, where $V$ is the set of vertices and $E$ is the set of edges. %and weighted graphs as $G=(V,E, W)$.
We represent the adjacency matrix as $A \in \{0,1\}^{n\times n }$, the number of vertices as $n$, the set of vertices as $V = \{1,\ldots,n\}$, and the set of edges as $E = \{(u,v) \in V \times V : A_{uv}=1\}$. For each node, we may consider a $d$-dimensional feature vector. We denote the feature matrix as $X \in \mathbb{R}^{n \times d}$. If no feature is associated with a node, $X$ is an identity matrix. Let us assume $C$ is the set of communities and the cardinality of $C$ is $k$. In the case of overlapping community detection, we assign each node to some of the communities in $C$. However, for each node, there is a strength associated with each different community. %communities. 
We %say 
define this strength %as 
as the community affiliation of the node. For each node, there is a $k$-dimensional community affiliation vector. We assign the community affiliation as a non-negative real number. Therefore, we can %get
obtain a community affiliation matrix as $F \in \mathbb{R}_{\geq 0}^{n \times k}$. For a node $u$, we denote the membership strength of community $c$ as $F_{uc}$. We use a threshold to obtain a binary community affiliation matrix as $F \in \{0,1\}^{n \times k}$. Therefore, each node can be assigned to multiple communities. It is also possible that some nodes have no communities at all. %In the
The following text %first
describes the basic concepts related to our paper. %, and then some related works.
% \subsection{Background}
%In the following, 
% Here, we %will 
% discuss some necessary concepts related to our paper.
% \\
% \subsubsection{Neural Network} 

\textit{Artificial neural networks} are %is
a class of machine learning models. The smallest computing unit of a neural network is a neuron. The input of a neuron is a vector. A neuron is a parameterized function whose output varies with different parameters for the same input. These neurons connected in a computational graph are called a neural network. For more information, readers may study the overview article by Schmidhuber \cite{schmidhuber2015deep}.
% \subsubsection{Convolutional Neural Network}

A \textit{convolutional neural network} (CNN) is a class of neural networks %which is 
based on convolution operation. We apply CNN on image structured data, which can be represented as an 8-regular graph   % It is possible consider an image as a regular graph, 
where each pixel is a node and each pixel is connected to eight neighboring 
%eight 
pixels. %Thus,  we may consider an image as an 8-regular graph. %Thus 
%Therefore, we may use a CNN for a regular graph. 
However, these classical CNNs are not able to process general graphs. %To perform convolution on general graphs we need special treatment
Special treatment is required to perform convolution on general graphs.

% \subsubsection{Graph Neural Network}
\textit{Graph neural networks} (GNN) is designed to handle the inability of classical neural networks (NN) to process general graphs. To alleviate this limitation of NN, Scarselli \textit{et. al.} proposed %the 
a graph neural network model \cite{scarselli2008graph}. % to alleviate the limitations of classical neural networks to process graph structured data. 
According to this model, a label vector and a feature vector are associated with each graph node. This model predicts the class of the graph or the classes of the nodes of the graph provided the labels and feature vectors for all nodes. 
% There are two types of functions %are 
% associated with this model: 1) transfer function and 2)  output function. The transfer function transforms the feature vector of each node to a new transformed vector based on the features of neighboring nodes. The transfer function is applied until the node %s 
% features come to a stable state, i.e., the features do %es 
% not change anymore. Finally, an output function is applied to %get 
% obtain the predicted class of each node. The output function can be node-wise or entire graph-wise. The %se 
% transfer function and output function are realized using a classical neural network. This is the basic idea of the very first graph neural network model. 

% \subsubsection{Graph Convolution Networks}
\textit{Graph convolutional networks} (GCN) are GNNs %which
that can perform convolution operations in general graphs. In the case of GCNs, each node aggregates the feature vectors of its neighboring nodes to achieve a richer feature vector. For each vertex $u$, a $d$-dimensional feature vector is considered as $h_u \in \mathbb{R}^d$. The whole graph $G$ can be represented by concatenating the feature vectors of all the nodes as $h_G=[h_{u_1},h_{u_2},\ldots,h_{u_n}]^T \in \mathbb{R}^{n \times d}$. %, where $N$ is the number of vertices. 
A general graph convolution operation is defined based on two operations: an aggregation operation and an update operation, 
\begin{align*}
    G_{l+1}= Update(Aggregate(G_l,W_l^{agg}),W_l^{update})
\end{align*}
where $G_l$ is the input graph with features at the $l$-th layer and $G_{l+1}$ is the output graph with updated features at the $(l+1)$-th layer. For the aggregation operation, the learnable weights matrix is $W^{agg}$ and for the update operation, the learnable weights matrix is $W^{update}$.
% In most  %the 
% graph neural networks, the aggregation function aggregates the features from the neighbors and the update function updates the feature vector of a node based on the output of the aggregation function. In most %of the 
% cases, the update function is a non-linear function. 
% Many variations of these functions are also possible. The mean aggregation function \cite{kipf2016semi}, max-pooling aggregation function \cite{wang2019dynamic,hamilton2017inductive,qi2017pointnet}, attention aggregation function \cite{velivckovic2017graph}, LSTM aggregation function \cite{peng2017cross}, etc., are some of the existing aggregation functions. MLP (multi-layer perceptrons) \cite{hamilton2017inductive, duvenaud2015convolutional}, gated network \cite{li2015gated}, etc., are some of the variations of update functions. 

\textit{Shallow GCNs} are GCNs having %very low
small numbers of layers ($<3$). In previous work, % by
NOCD %\cite{shchur2019overlapping} %, the authors 
used a two-layered %graph convolution network
shallow GCN that was %is
first proposed by Kipf and Welling \cite{kipf2016semi}. %We refer to this two-layer GCN as a shallow GCN. 
According to NOCD, %\cite{shchur2019overlapping},
the GCN model is defined as follows:
\begin{align*}
    F:=GCN_{\theta}(A,X)=ReLU(\hat{A}(ReLU(\hat{A}XW^{(1)}))W^{(2)})
\end{align*}
Here, $\hat{A}=\tilde{D}^{-1/2}\tilde{A}\tilde{D}^{-1/2}$ is the normalized adjacency matrix, and $\tilde{A}=A+I_N$ is the adjacency matrix considering self-loops. The diagonal degree matrix of $\tilde{A}$ is $\tilde{D}_{ii}=\sum_j{\tilde{A}_{ij}}$. Here, the $ReLU$ function works as a non-linear update function and aggregation is done through a weighted sum of the features of neighbors. This weight is a learnable parameter. 

% \remove{According to this model, deep architectures do not work well \cite{kipf2016semi, shchur2019overlapping}. %In this shallow model some 
% Some modifications to this shallow model were introduced \cite{shchur2019overlapping}. These are: (1) batch normalization after the first GCN layer and (2) $L_2$ regularization applied to all weight matrices. In fact, shallow GCNs have some limitations. %In GCN, t
% For instance, the number of layers is limited. When the number of layers %are
% is increased, a GCN faces the over-smoothing and the vanishing gradient problem.}

\subsection{Overlapping Community Detection Methods}
In the following, we include most of the prominent related works which we have used for comparison and evaluation.

\subsubsection{SLPA}
The Speaker-listener Label Propagation Algorithm (SLPA) \cite{xie2011slpa} is a general framework to analyze overlapping communities in social networks. According to this algorithm, nodes exchange information (label) based on dynamic interaction rules. This framework is designed to analyze both individual overlapping nodes and the whole community. SLPA is an extension of the previous label propagation algorithm (LPA) \cite{raghavan2007nearlpa}. %In label propagation algorithm, e
Each node in the LPA has a single label. This label is %interactively
iteratively updated by the majority of labels in the neighborhood. After completion of the algorithm, non-overlapping (disjoint) communities are discovered. To allow overlap, each node is allowed to have multiple labels. 
% \remove{There are two issues related to %In 
% this algorithm: %there are two issues:
% 1) disseminating node information efficiently and, 2) aggregating/processing information received from neighbors in an information preserving way. Another critical issue is %with 
% related to information maintenance. %To resolve all of these issues and manage data at different nodes, 
% The authors of SLPA designed a speaker-listener based algorithm to resolve all of these issues and manage data at different nodes. In the SLPA, each node can act as a listener or a speaker, depending on whether it is absorbing information or disseminating information. The algorithm consists of three stages. First, each node label is initialized with a unique value. Then, an iterative process continues until a stopping condition is met. 
% % The process is as follows \cite{raghavan2007nearlpa}:
% % \begin{itemize}
% %     \item A node is considered as a listener
% %     \item Each neighboring node sends a label to the node, according to some predefined \textit{speaking} rule
% %     \item The listener accepts one label from all of the labels received, according to a predefined \textit{listening rule}
% % \end{itemize}
% Finally, post-processing of the accumulated labels %of
% for each node determines the communities. 
% The algorithm runs in $\mathcal{O}(Tn)$. Here, $n$ is the number of nodes in the graph and $T$ is the maximum number of iterations. 
% }

\subsubsection{BigClam} BigClam (Cluster Affiliation Model for Big Networks) was originally designed for large networks \cite{yang2013overlapping-bigclam}. BigClam is a probabilistic generative model for graphs to capture network community structure based on the community distribution of each node. The whole idea of BigClam is based on a bipartite community affiliation network where 
the generative model is based on the fact that the higher number of communities are shared by two nodes, %share, 
the greater %is 
the probability of connecting these two nodes with an edge. 
Community detection using the BigClam model is the reverse problem of generating the graph. The community affiliation matrix $F$ is determined by the underlying graph $G(V, E)$. The number of communities $k$ is given, and the BigClam model finds the most likely affiliation matrix $\hat{F}$ maximizing log-likelihood. %}

% $l(F) = \log{P(G|F)}$. 
% \begin{align}
%     \hat{F} &= \argmax_{F\geq0} l(F)
% \end{align}
% \begin{align}
%     l(F) &= \sum_{(u,v)\in E}\log{(1-exp(-F_{u}F_{v}^{T}))}- \sum_{(u,v)\not\in E} {F_uF_v^T}.
% \end{align}
% This optimization is similar to the non-negative matrix factorization approach (NMF) \cite{lee1999learning}. %Optimization goal 
% The goal of the optimization is to learn the affiliation matrix $F$ that best approximates the underlying graph $G(V,E)$. %best. 
% Existing NMF methods \cite{hsieh2011fast,lee1999learning} %are modified by 
% have been modified by BigClam for large networks \cite{yang2013overlapping-bigclam}. BigClam %adopts adopted 
% a block co-ordinate ascent algorithm \cite{hsieh2011fast,lin2007projected} to solve the optimization problem. After finally %getting 
% obtaining the community affiliation matrix $F$, a threshold $\delta$ is used to determine the community affiliation of each node. If $F_{uc}< \delta$, then node $u$ does not belong to the community $c$. 

\subsubsection{DEMON}
DEMON stands for Democratic Estimate of the Modular Organization of a Network \cite{coscia2014uncovering-demon}, %. In this approach, the authors designed 
an algorithm designed to discover communities in complex networks. In this approach, each node gives community labels to its neighboring nodes. Thus, every node %gets 
obtains community labels from its neighboring nodes. This approach is called a \textit{democratic approach} because each node can vote for every neighboring node. %s. 
Then, the communities are discovered from the labels for each node. %the communities are discovered. 
Since one node can have multiple labels, one node can belong to multiple communities. After obtaining labels from the neighborhood, the communities can be discovered using different merging algorithms. 
% \remove{For each node, its corresponding ego network is extracted. Then, a label propagation community detection algorithm \cite{raghavan2007nearlpa} is applied to remove %removing 
% the corresponding node. Finally, a combination step is executed to %get 
% obtain the communities. }

\subsubsection{MNMF}
Wang et al. \cite{wang2017community-mnmf} proposed the Modularized Nonnegative Matrix Factorization (M-NMF) model. In this approach, the authors designed a community-preserving matrix factorization scheme. The modularity-based community detection model and NMF-based representation learning model are jointly optimized in a unified learning framework. Basic intuition %is 
indicates the nodes belonging to the same community should have similar representations. The proposed M-NMF model is optimized to preserve both pairwise node similarity and community structure. For pairwise node similarity, the first-order and second-order proximities of nodes are incorporated to learn representations using matrix factorization. A modularity constraint term is incorporated to detect communities.

\subsubsection{DANMF}
% \remove{Non-negative matrix factorization (NMF) is a prominent approach for community detection \cite{wang2011community-snmf,yang2013overlapping-bigclam,kuang2012symmetric}. NMF approaches have better interpretability and %it is 
% are a natural %ly 
% fit for both overlapping community detection and disjoint (non-overlapping) community detection.} 
DANMF \cite{ye2018deep-danmf} (Deep Auto-encoder like Non-negative Matrix Factorization) is a variant of NMF. NMF-based approaches factorize %s 
the adjacency matrix $A$ into two non-negative matrices $U$, $V$, where $A\sim UV  (U\geq 0, V\geq 0)$. Normally, the columns of $V$ represent the community strength of the nodes of the network. %Normally 
%NMF methods typically follow shallow methods, i.e., the adjacency matrix is factorized in one level.
% \newpage
% \begin{figure}[h]
%     \centering
%     \includegraphics[width=0.5\textwidth]{images/danmf/nmf-deepnmf.png}
%     \caption{ NMF and deep NMF\cite{ye2018deep-danmf}) }
%     \label{fig:nmf1}
% \end{figure}
The authors  of DANMF proposed a hierarchical matrix factorization scheme. Inspired by deep auto-encoders, the adjacency matrix %will be
is factorized into multiple levels.
%, as shown in figure \ref{fig:nmf2}. 
% \begin{figure}[h]
%     \centering
%     \includegraphics[width=0.5\textwidth]{images/danmf/deepDANMF.png}
%     \caption{ deep NMF architecture\cite{ye2018deep-danmf}) }
%     \label{fig:nmf2}
% \end{figure}
% \newpage
DANMF is like auto-encoders where there is an encoder component and a decoder component. Both components %are having 
have deep structures. The encoder part transforms the original adjacency matrix to a community affiliation matrix and the decoder part reconstructs the original adjacency matrix. Unlike %deep-autoenoder
deep-autoencoders, the loss function in DANMF %loss function 
is unified with both the encoder and decoder. 
% \remove{Thus, DANMF is similar to deep auto-encoders for representation learning\cite{bengio2013representation}.}

% \section{Literature Review:} Community detection in graphs is a well known problem and research topic. However, most of the works in the literature are related to non-overlapping community detection \cite{abbe2017community,von2007tutorial}.    None of these approaches considered overlapping community detection.   Jia \textit{et. al.} \cite{jia2019communitygan} designed generative adversarial nets, named communityGAN, to perform community detection.  In our work, we used deep graph convolutional neural networks.

\subsubsection{NOCD}
NOCD  stands for neural overlapping community detection \cite{shchur2019overlapping}. The model consists of an encoder and a decoder. The encoder is a shallow graph convolution network (GCN) with two layers. The decoder is based on the Bernoulli-Poisson model \cite{yang2013overlapping-bigclam}. The output of the encoder is a community affiliation matrix. The decoder attempts to reconstruct the original graph. Finally, a loss is generated based on the reconstruction error. This loss is used to train the GCN model. We use a similar framework in this study.  

We also compare our method with some other related works including EPM \cite{zhou2015infinite-epm}, SNetOC \cite{todeschini2016exchangeable-snetoc}, CESNA \cite{yang2013community-cesna}, SNMF \cite{wang2011community-snmf}, CDE \cite{li2018community-cde}, DeepWalk \cite{perozzi2014deepwalk} with Non-exhaustive Ovelapping K-means \cite{whang2015non-neo}(DW/NEO) , and Graph2Gauss \cite{bojchevski2017deep-g2g} with Non-exhaustive Overlapping K-means \cite{whang2015non-neo} (G2G/NEO).

\section{Our Methodology}

% We use the same notations as in \cite{shchur2019overlapping}. We consider undirected unweighted graph as $G=(V,E)$ where $V$ is the set of vertices and $E$ is the set of edges. We represent the adjacency matrix as $A \epsilon \{0,1\}^{N\times N }$. Number of nodes are represented as $N$ and the vertices as $V = \{1,\ldots,N\}$. Number of edges are represented as $M$ and edges as $E = \{(u,v) \epsilon V \times V : A_{uv}=1\}$. For each node, we can consider a $D$-dimensional feature vector. The feature matrix is represented as $X \epsilon \mathbb{R}^{N \times D}$. Let's assume $C$ is the number of communities. In case of overlapping community detection, we assign each node into $C$ communities. However, for each node, there is a strength associated with each different communities. We say this strength as the community affiliation. For each node, there is a $C$-dimensional community affiliation vector. The community affiliation or strength is assigned as a non-negative real number. Therefore, we can get a community affiliation matrix as $F \epsilon \mathbb{R}_{\geq 0}^{N \times C}$. For a node $u$, the membership strength in community $c$ is denoted as $F_{uc}$. Sometimes, we use a threshold to get a binary community affiliation matrix as $F \epsilon \{0,1\}^{N \times C}$. Therefore, each node can be assigned to multiple communities. It is also possible that some nodes have no communities at all. 
Communities are defined on a graph as the dense sub-graphs of the graph. According to the most %of the 
recent research, %ers, 
a community is a group of nodes. %For
These nodes have a higher probability of forming %, the probability to form 
 edges within the group %is greater 
than forming edges with %the 
nodes outside of the group \cite{fortunato2016community,shchur2019overlapping}. We adopt this definition of a community to develop our framework of overlapping community detection. We consider a graph generative model $p(G|F)$ for a graph $G$ \cite{yang2013overlapping-bigclam,shchur2019overlapping}. 
% Here, $F$ is the community affiliation matrix. %This is a $N\times C$ matrix, where $N$ is the number nodes in the graph and $C$ is the number of communities in the graph. 
% $F_{ij}$ represents the strength of node $i$ in community $j$, i.e., the affiliation of node $i$ to community $j$.  
According to this model $p(G|F)$, we can generate the graph $G$ for a given $F$. However, overlapping community detection is exactly the opposite task. We have to find the unobserved affiliation matrix $F$ for a given %a
graph $G$ \cite{yang2013overlapping-bigclam,shchur2019overlapping}. We can optimize $F$ by minimizing some loss function based on the model $p(G|F)$ \cite{yang2013overlapping-bigclam}. On the contrary, we can feed the graph $G$ into a GNN; then we can optimize the GNN to generate a more accurate $F$. %\cite{shchur2019overlapping}. The %later 
%latter approach performs better in most %of the 
%cases \cite{shchur2019overlapping}. 
In this paper, we follow the GNN-based approach to optimize $F$.

\begin{figure}[h]
    \centering
    \includegraphics[width=0.8\textwidth]{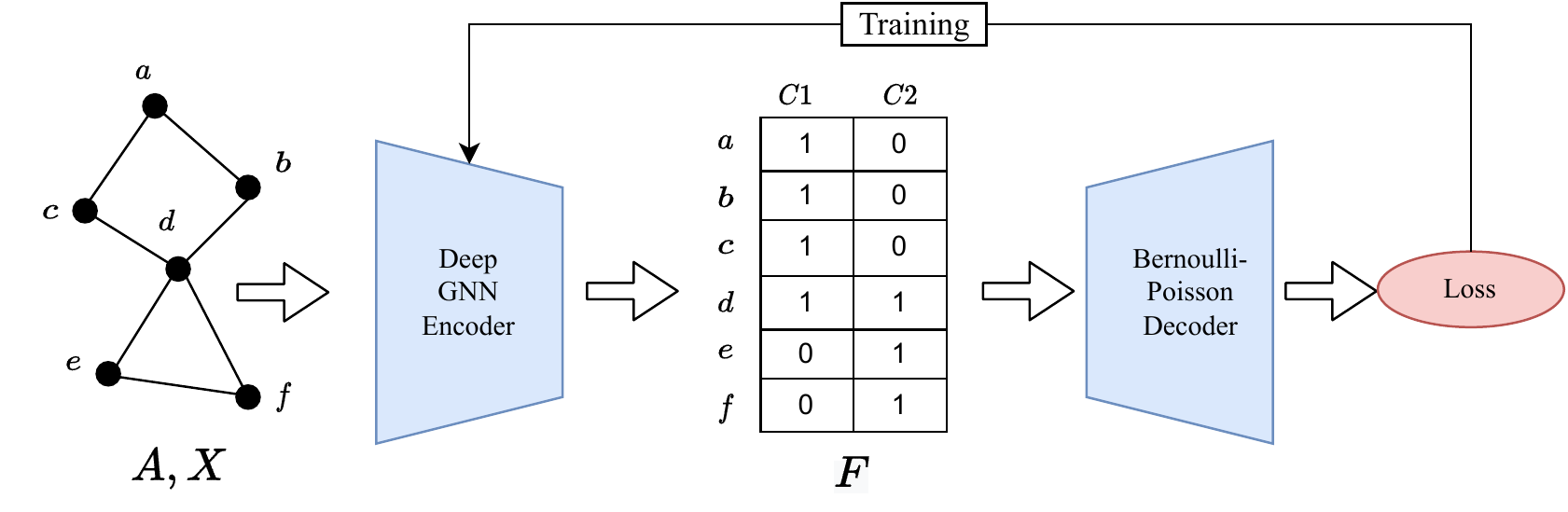}
    \caption{The overlapping community detection model %is 
    based on the deep DynaResGCN encoder and the Bernoulli-Poisson decoder.}
    \label{fig:model}
\end{figure}

%In this study, 
We consider an unsupervised approach in this study. Essentially, we design an encoder and decoder-based approach to tackle the problem of overlapping community detection (Fig. \ref{fig:model}). The encoder takes %took 
a graph %$G$
with adjacency matrix $A$ and feature matrix $X$ as input and generates the community affiliation matrix $F$ as output. The decoder takes %took the 
$F$ as input and produces %d 
a loss. We use the loss generated by the decoder  to train the encoder. Indeed, we use a deep GCN-based encoder %\cite{shchur2019overlapping, kipf2016semi} % As a decoder, 
and the Bernoulli-Poisson decoder. % \cite{shchur2019overlapping}. 
%To the best of our knowledge
To achieve deep GCNs, we %are the first to 
introduce the concept of dynamic dilated aggregation successfully in the case of %general 
irregular graphs. % to achieve deep GCNs. \textcolor{red}{
We also incorporate the idea of residual connection %, and dynamic edge 
in our model. We term our deep GCN model DynaResGCN. %Moreover, we explore different approaches to consider edge-weights of weighted graphs. These approaches combined with DynaResGCN are described in Table \ref{tab:my-methods}. 
In a nutshell, we unify the idea of dynamic edges, dilated aggregation, and residual connection %and consideration of edge-weights 
in an encoder-decoder-based framework for overlapping community detection. Next, we describe our approach to achieving the Deep GCN (DynaResGCN) encoder, the corresponding Bernoulli-Poisson decoder to train the Deep-GCN encoder, and the training algorithm.

\subsection{Deep Dynamic Residual GCN Encoder}
\begin{figure}[h]
    \centering
    \includegraphics[width=0.95\textwidth]{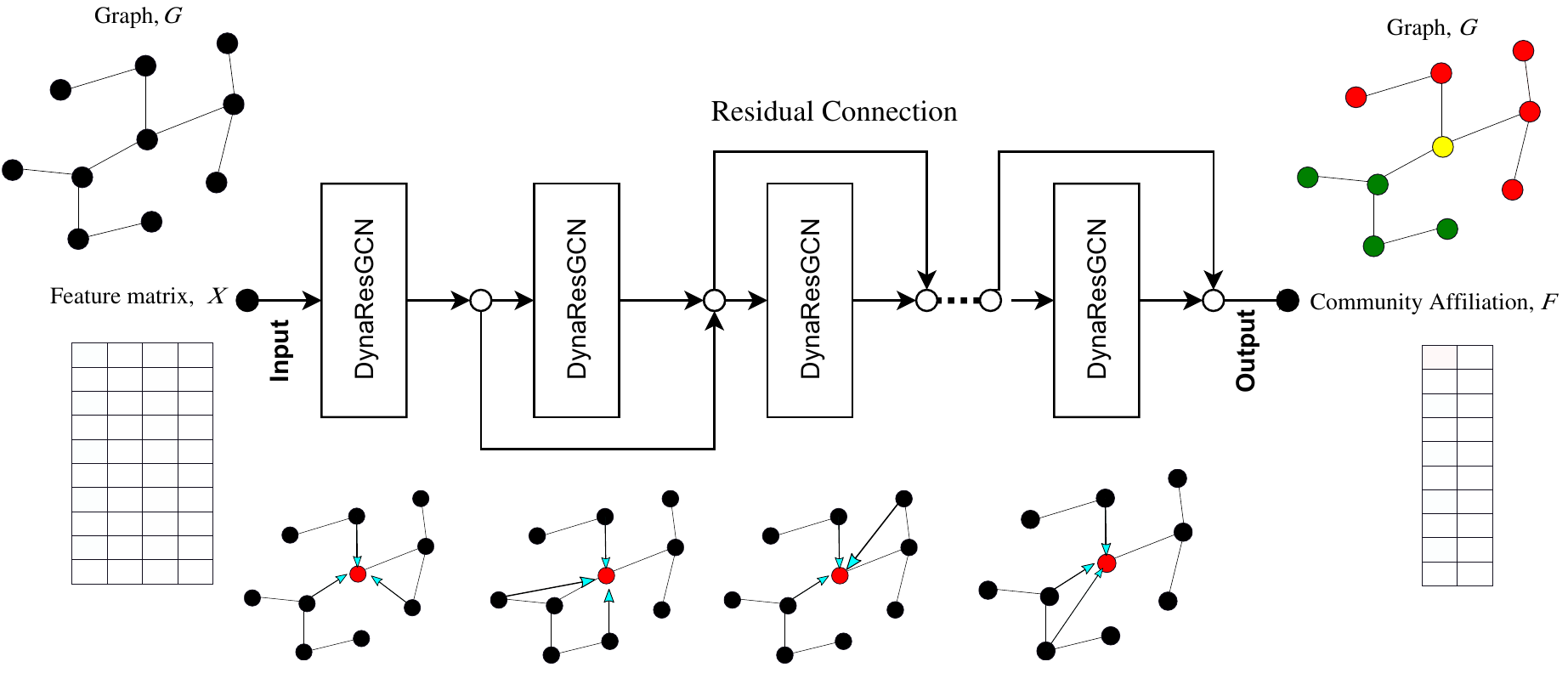}
    \caption{%This figure explains the whole architecture of the Deep DynaResGCN part. We show message passing of GNN for only a single node (for simplicity). At each layer, information is aggregated from a fixed number of random nodes. Some of these nodes are chosen from the first hop and others are chosen from the second hop neighbors based on the  dynamic dilated aggregation algorithm. %Thus dynamcity of neighborhood is incorporated.
    The architecture of the deep DynaResGCN encoder.
    }
    \label{fig:architecture}
\end{figure}

A deep-GCN encoder is a GNN with many layers %that has
having the capability of effective learning. 
% \remove{However, it is not possible to achieve a deep-GCN by increasing the number of layers in shallow models due to the over-smoothing and vanishing gradient problems \cite{li2019deepgcns}.}
In this study, we introduce a deep-GCN encoder for overlapping community detection in %general irregular
graphs. We %unify
incorporate the idea of residual connections, dynamic edges, and dilated aggregation %, and edge-weights
in a single framework called DynaResGCN (Fig. \ref{fig:architecture}). This figure explains the whole architecture of the deep DynaResGCN encoder where message passing of GNN is shown for only a single node (for simplicity). At each layer, information is aggregated from a fixed number of random nodes. Some of these nodes are chosen from the first hop and others are chosen from the second hop neighbors based on the  dynamic dilated aggregation algorithm. The descriptions of each part of the architecture are provided in the following text.
% \remove{Incorporating these concepts in the case of general graphs is challenging because of the variable neighborhood size of general graphs.}
%The approaches to employing these ideas %in general irregular graphs
%, and further unification, 
%are described below:

\subsubsection{Residual Connection} 
% \remove{The idea of residual connections was first introduced in CNN for image classification \cite{he2016deepResNet}. Later, this idea was transferred into GCN for point cloud learning \cite{li2019deepgcns} and has been successfully applied to achieve deep models. In fact, residual connections solve the vanishing gradient problem \cite{he2016deepResNet}.}
In this study, we incorporate residual connections into the GCN encoder for overlapping community detection. If $\hat{A}$ is the normalized adjacency matrix, %\cite{shchur2019overlapping}, 
$X_l$ and $W_l$ are the feature matrix %is $X_l$ at layer $l$  and the 
and learnable weight matrix %is $W_l$
at layer $l$ respectively, then computation of the GCN encoder at layer $l$ is expressed as:
\begin{align*}
    X_{l+1} &= ReLU(\hat{A}X_lW_{l})
\end{align*}
In our study, we incorporated residual connection at layer $l$, as follows:
\begin{align*}
    X_{l+1} &= ReLU(\hat{A}X_lW_{l})+\hat{A}X_l \numberthis \label{eq1} \\
            &= X_{l+1}^{res} + \hat{A}X_l
\end{align*}
The additional term in equation \eqref{eq1} is $\hat{A}X_l$. This term connects the input $X_l$ with the output $X_{l+1}$. Consequently, a skip connection is created in the computation path, which enables a smooth flow of gradients %and therefore, gradients do not vanish away, which 
%potentially
for solving the vanishing gradient problem. 

\subsubsection{Dilated Aggregation}
% \remove{The concept of dilation came from the wavelet processing domain. Specifically, the  algorithm named \emph{dilated wavelet convolution} was designed for wavelet processing \cite{holschneider1990real, shensa1992discrete}. For predictive learning, Yu and Koltun \cite{yu2015multiDynamicEdge} proposed dilated convolution for dense prediction tasks. Their aim was to compensate for spatial information loss due to pooling operations. Dilated convolutions were shown to significantly improve the accuracy of dense prediction tasks through aggregating multi-scale contextual information \cite{yu2015multiDynamicEdge}. In fact, dilation increases the receptive field without  leading to a loss of resolution \cite{li2019deepgcns}. Consequently,} 
Dilated aggregation was applied to GNNs for point cloud learning \cite{li2019deepgcns}. In the point cloud, a neighborhood is constructed for each point using the k-nearest neighbor (k-NN) approach \cite{li2019deepgcns}. Then, dilation is applied through sub-sampling from the fixed neighborhood \cite{li2019deepgcns}. For instance, %let us assume that
if $k=10$, then the nearest neighborhood of size $10$ is constructed from the point cloud. The nearest neighbors can be ordered for each point based on the distance. To perform dilation, every even-numbered neighbor can be skipped, and every odd-numbered neighbor can be considered. Thus, aggregation for information could be done from the first, third, fifth, seventh, and ninth neighbors. %As a result,
Therefore, this aggregation is called dilated aggregation. Using dilated aggregation, it is possible to reach distant neighbors without increasing the number of neighbors considered for aggregation.

However, there is no flexibility in the neighborhood size in general graphs. For each node, the neighborhood is defined according to the graph structure. In fact, the size of the neighborhood  in general graphs varies widely. Therefore, the fixed-sized k-NN approaches are not applicable. % to the general graphs. 
An adaptive approach that accounts for the variable neighborhood size must be designed. It is also possible to define the neighborhood of the general graphs based on proximity. For instance, we can define first-order proximity based on the first-hop neighbors of a node. Accordingly, second-order proximity may be defined considering all first- and second-hop neighbors. If we consider the first-order proximity and sub-sample from the first-hop neighbors, there is a possibility of information loss due to losing a number of first-hop neighbors. In the case of higher-order proximity, the neighborhood size can be large and intractable in practical cases. Moreover, a large number of neighbors can boost over-smoothing. Over-smoothing means very similar features for all the nodes in such a way that all the nodes lose distinguishability. To tackle these challenges, we designed %novel dynamic 
dilated aggregation algorithm, which can be combined with the concepts of dynamic edges.% and edge-weights. 

In our design, we have to handle two rival issues. Firstly, dilation in first-order proximity results in information loss. On the contrary, the neighborhood size increases at a quadratic rate when second-order proximity is considered. To achieve a fair balance between these rival issues, we choose the augmented neighborhood size of a node as $2*m$ where $m$ is the degree of the node. Then, we choose $m$ nodes from the first hop and the remaining $m$ nodes from the second hop. We design an augmentation algorithm (Algorithm \ref{alg:random-augmentation}) %, Algorithm \ref{alg:weighted-augmentation}) 
to choose the nodes from the second-hop neighbors. %Using weighted augmentation algorithm, we can exploit the information in the weights of the edges to choose neighbors from the second hop. 
Finally, we introduce dilation in this augmented neighborhood. We have designed a simple but effective dilated aggregation scheme. %s where we can exploit information coming from the edge weights. 
The dilation scheme is to sub-sample $50\%$ nodes randomly from the augmented neighborhood. After sub-sampling, each node would have %\textcolor{red}{
$m$ neighbors effectively. Thus, we do not increase the effective neighborhood size while we can aggregate information from the distant nodes.

Let us consider an $m$-regular graph. In this case, first-order proximity considers a neighborhood of size $m$. However, the second-order proximity would consider $m^2+m$ neighbors. Our approach considers an augmented neighborhood of size $2*m$, which is not much larger. To implement dilation, we sample $50\%$ nodes from this augmented neighborhood. Therefore, the effective number of neighbors is $2*m*0.5=m$, which is the same as the number of original first-hop neighbors. Therefore, the possibility of information loss due to dilation is diminished with the advantage of a larger receptive field.

% Now, the question arises that how to achieve the augmented neighborhood. It is possible to choose $d$ nodes from the first-hop neighbors. However, the issue is, how to choose the remaining $2*d-d=d$ nodes from the second-hop neighborhood of size $d^2$. To resolve this issue, we designed augmentation algorithms, which are explained later. Finally, to implement dilation, we designed different sampling procedures based on various criteria.  

\subsubsection{Dynamicity of Edges} The dilated aggregation procedures inherently lose some first-hop neighbors. When the same dilated neighborhood is considered at every GNN layer, some first-hop neighbors are never explored for each node. This fixed neighborhood at every layer effectively alters the original graph. Moreover, this introduces over-smoothing as the information from the same neighbors is aggregated at each layer. Consequently, learning is not possible. To remedy this issue, we consider different sub-samples of neighbors from the augmented neighborhood at every different layer. Therefore, the neighborhood is different at every layer. In essence, the edge set of the graph becomes dynamic. Interestingly, the concept of dynamic edges and dilated aggregation can be unified. As a result, learning becomes stable and better than static approaches  \cite{wang2019dynamic,valsesia2018learning,simonovsky2017dynamic}. In the case of point clouds, the nearest neighbor graph is dynamically constructed after every layer  \cite{wang2019dynamic,valsesia2018learning}. To introduce this concept into the DynaResGCN encoder, the adjacency matrix is constructed differently at each layer based on the augmentation algorithm and dilation scheme. Let us assume $A_l$ is the adjacency matrix at layer $l$ where $A_l$ differs in different layers. Therefore, the computation at layer $l$ defined in \eqref{eq1} is updated to  the following 
\begin{align*}
    X_{l+1} &= ReLU(\hat{A_l}X_lW_{l})+\hat{A_l}X_l \numberthis \label{eq2}\\
            &= X_{l+1}^{res} + \hat{A_l}X_l
\end{align*}

\begin{algorithm}[t] 
%   \scriptsize
    \caption{\bfseries{Graph Augmentation: Input $G=(V,E)$}}
   
    \begin{algorithmic}[1]
    \State Let $E^\prime = E$
    \For {each node $u$ in the graph}
    % \State Let $d$ be the degree of node $u$
    % \State $k= d \times 2 $
    \State Let $S_u$ be the set of original neighbors of $u$
    \State Let $S^\prime_u$ be the set of augmented neighbors of $u$
    \State Initialize $S^\prime_u$ with the  elements of $S_u$
    \For{ each node $v\in S_u$}
    \State Obtain a random node $w \not \in S^\prime_u$ from the neighbors of $v$  ($w\in S_v$) %which are not neighbors of $u$
    \State Add $w$ to $S^\prime_u$
    \State Add edge $(u,w)$ to $E^\prime$
    \EndFor
    \EndFor
    \State Return  $G^\prime=(V,E^\prime)$ considering augmented neighborhood
   
    \end{algorithmic}
    \label{alg:random-augmentation}
\end{algorithm}

% \begin{algorithm}[H] 
% %   \ref{alg:augmentation}
%     % \scriptsize
%     \caption{\bfseries{Graph Augmentation (Weighted)}}
   
%     \begin{algorithmic}[1]
%     \For{ each node $u$ in the graph}
%     % \State Let $d$ be the degree of node $u$
%     % \State $k= d \times 2 $
%     \State Let $S$ be the set of neighbors of $u$
    
%     \For{ each node $v\in S$}
%     \State obtain the most edge-weighted node $w$ from all neighbors of $v$ which are not neighbors of $u$
%     \State set $w$ as a new neighbor of $u$

%     \EndFor
%     \EndFor
   
%     \end{algorithmic}
%     \label{alg:weighted-augmentation}
% \end{algorithm}

\subsubsection{Dynamic Dilated Aggregation}
The ideas of residual connection, dilated aggregation, and dynamic edges are unified in a single framework by the equation \eqref{eq2}. This section defines the augmentation algorithm and %different 
dilated aggregation techniques. The adjacency matrices $\hat{A_l}$ are generated using these algorithms. %We employ two augmentation algorithms: 1) random augmentation (Algorithm \ref{alg:random-augmentation}), and 2) weighted augmentation (Algorithm \ref{alg:weighted-augmentation}). 
Our augmentation algorithm is described in the Algorithm
\ref{alg:random-augmentation}. 
\begin{figure}[h]
    \centering
    \includegraphics[width=0.5\textwidth]{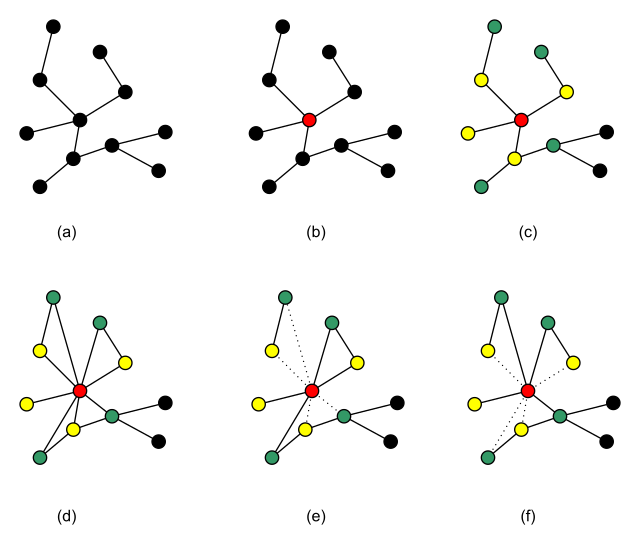}
    \caption{(a) %Shows
    A graph (network) (b) in which a node (red) is considered, (c) first-hop neighbors and second-hop neighbors are indicated in yellow and green, respectively, (d) graph augmentation up to second-hop neighbors, (e) dilated %filtering
    neighborhood at one layer (random), and (f) dilated %filtering
    neighborhood at another layer (random) }
    \label{fig:nbragg2}
\end{figure}
The random augmentation algorithm is applicable to any graph. %However, the weighted augmentation algorithm exploits the information coming from the edge-weights of a weighted graph.
In Figure \ref{fig:nbragg2}, we show %random
graph augmentation and random sub-sampling of neighbors for a single node. 
In the following text, we describe a dilated aggregation scheme based on the %different 
sub-sampling/dilation 
%procedure
of an augmented neighborhood.

\par
\textbf{Random Dynamic Dilated Aggregation:}
For random dynamic dilated aggregation, we first apply  Algorithm \ref{alg:random-augmentation}. All the first-hop neighbors are considered in this algorithm. Additionally, for each first-hop neighbor, a new node from the second-hop neighborhood, connected to the first-hop node, is considered. Now, we randomly sample 50\% of the nodes from the augmented neighbors at each layer. The sampling at each layer differs from other layers, allowing for exploring different neighborhoods. Thus, we employ the concept of dynamic edges in terms of dynamic neighborhoods with dilated aggregation.  This random dynamic dilated aggregation helps resolve the over-smoothing problem \cite{li2019deepgcns}. %However, the question is, how does this random dilated aggregation contribute to resolve over-smoothing problem? 

% \par
% \textbf{Weighted Dynamic Dilated Aggregation:} For weighted dynamic dilated aggregation, we first apply Algorithm \ref{alg:weighted-augmentation}. All of the first-hop neighbors are considered in this algorithm. Additionally, for each first-hop neighbor, a new node is considered from the second-hop neighborhood connected to the first-hop node. This new node is the most weighted node connected to the first-hop node. The remainder of this process is similar to random dynamic dilated aggregation. In this aggregation scheme, we attempt to exploit the information coming from edge-weights.
% \par
% \textbf{Weighted Differential Dilated Aggregation:} In this approach, a weighted augmentation algorithm (Algorithm \ref{alg:weighted-augmentation}) is used to augment the graph. Then at each layer, importance sampling is performed, in which the first-hop neighbors are given more importance than the second-hop neighbors. However, the amount of  priority to be given to the first-hop neighbors over the second-hop neighbors is a hyper-parameter. In our experiment, we consider 80\% importance for the first-hop neighbors and 20\% importance for the second-hop neighbors. As a result, the first-hop neighbors are selected with a probability of $0.80$, and the second-hop neighbors are selected with a probability of $0.20$. The intuition behind this step is that if a neighbor is more important to  a node, then it should be directly connected to the node. Being a second-hop neighbor automatically implies less importance, irrespective of the weight.

\subsection{Bernoulli-Poisson Decoder}
The final layer of the GCN encoder generates the affiliation matrix, $F$. To train the GCN, we need to calculate the loss. To generate the loss, we exploit the Bernoulli-Poisson model. %In fact, the BigClam model \cite{yang2013overlapping-bigclam} and other works \cite{zhou2015infinite-epm,todeschini2016exchangeable-snetoc} proposed the Bernoulli-Poisson (BP) model. The BP model is a graph generative model. 
According to this model, for each node $u$, there is a community affiliation vector $F_u$ with dimension $k$. 
% Here, $k$ is the number of communities. Considering $N$ nodes, the affiliation matrix is $F \in \mathbb{R}_{\geq 0}^{N \times C}$. 
This model generates the adjacency matrix $A$, for a given community affiliation matrix $F$, as follows:
\begin{align*}
    A_{uv} \sim \textrm{Bernoulli}(1-exp(-F_uF_v^T))
\end{align*}
% For a community $c$ and nodes $u$ and $v$, there is a probability of forming edge between $u$ and  $v$. This probability is related to the affiliation of nodes $u$ and  $v$ to community $c$. That probability is $1-exp(-F_{uc}.F_{uv})$. Each community has an independent probability of forming edge between every pair of nodes in the graph. In essence, the probability of an edge between $u$ and $v$, considering every community, is given by $1-exp(-\sum_c F_{uc}.F_{vc})$ \cite{yang2013overlapping-bigclam}.
% This BP model can generate nested and hierarchical communities and also can model densely overlapping communities \cite{yang2014structure, shchur2019overlapping}. Yang \emph{et al.} \cite{yang2013overlapping-bigclam} performed maximum likelihood estimation (MLE) to infer communities in the BP model with coordinate ascent.   In contrast, Zhou \emph{et al.} \cite{zhou2015infinite-epm} applied a Markov chain Monte Carlo estimation to infer overlapping communities. For our case, we assume $F$ is generated by the GNN encoder,
For a given graph with adjacency matrix $A$ and feature matrix $X$, we generate the community affiliation matrix using the DynaResGCN encoder as follows:
\begin{align*}
    F &= \textrm{DynaResGCN}_{\theta}(A,X)
\end{align*}
Now, according to the BP model, the negative log-likelihood is \cite{shchur2019overlapping}
\begin{align*}
    -\log {p(A|F)} &= -\sum_{(u,v) \in E} \log (1-exp(1-F_uF_v^T))+ \sum_{(u,v) \not\in E } F_uF_v^T.
\end{align*}
Due to the sparsity of many real-life graphs, non-edges are much more common than  edges. To overcome this issue, %the authors of 
NOCD %\cite{shchur2019overlapping} 
proposed  balancing the terms of  the negative log-likelihood 
according to the standard technique for imbalanced data
\cite{he2009learning} as  
\begin{align*}
    \mathcal{L}(F) &= - \mathbb{E}_{(u,v)\sim P_E}[\log(1-exp(-F_uF_v^T))]+ \mathbb{E}_{(u,v) \sim P_N}[F_uF_v^T], \numberthis \label{eq3}
\end{align*}
where uniform distributions over edges and non-edges are represented by $P_E$ and $P_N$, respectively. In our case, we follow the same approach. %as in NOCD \cite{shchur2019overlapping}. 
To learn the community affiliation matrix $F$, the authors of BigClam %\cite{yang2013community-cesna, yang2013overlapping-bigclam} 
directly optimized the affiliation matrix $F$. Unlike these traditional approaches, we optimize the parameters of the GNN to obtain a more accurate affiliation matrix. The equation \eqref{eq3} generates the loss. This loss is back-propagated in the GNN encoder to learn the parameters $W_l$. Next, we describe our training algorithm.
\subsection{Training Algorithm}
We consider $\tau$ as the total number of epochs for simplicity. In our implementation, we consider the early stopping criterion when the learning becomes stable. Our training algorithm, Algorithm \ref{alg:training-algorithm}, is as follows:
\begin{algorithm}[H]
%   \scriptsize
    \caption{\bfseries{Training Algorithm}}
   
    \begin{algorithmic}[1]
    \State Perform graph augmentation (random)%, weighted) 
    \State Build the GNN encoder with a predefined number of layers $n$
    \State For each layer, generate %the adjacency matrix 
    $\hat{A_l}$ based on the dilated aggregation scheme
    \For {each epoch $e$ in $\tau$}
    % \State For stochastic learning, sample a batch edges $E$.
    % \State Feed the normalized feature matrix $X$ to the GNN as follows
    \For {each layer $l$ in total $p$ layers}
       \State $X_{l+1} = \textrm{Relu}(\hat{A}_lX_lW_l)+ \hat{A}_lX_l $
    \EndFor
    \State Get %the community affiliation matrix 
    $F=X_p$ from the output of the GNN
    \State Calculate loss and update parameters
    % using equation \eqref{eq3}
    % \State Calculate gradients using this loss
    % \State Update the network parameters $W_l$ using the gradients
    
    \EndFor
    \end{algorithmic}
    \label{alg:training-algorithm}
\end{algorithm}
\section{Experimentation}
In this study, %we propose different methods based on the different dilated aggregation schemes and different types of graphs (weighted/unweighted). These methods are listed in the Table \ref{tab:my-methods}. In order to perform evaluation and validation, 
we have extended the code base of NOCD %\cite{shchur2019overlapping} 
to implement our ideas. We have two types of datasets: 1) without ground truth and 2) with ground truth. The research topics dataset ($n \approx 6K $) \cite{burd2018gram} is a medium-sized dataset without ground truth information. All other datasets have ground-truth information \cite{mcauley2014discovering-facebook-dataset,shchur2019overlapping}. These datasets with ground truth are of two types: 1) small ($n < 800$) datasets (Facebook datasets \cite{mcauley2014discovering-facebook-dataset}), and 2) very large ($n> 20K$ up to $65K$) datasets (\cite{shchur2019overlapping}). %For hyper-parameter optimization, we use grid search. 
We vary the number of layers from 10 to 90 in steps of 10 for the topics dataset. %in the case of every different methods we have designed.
However, for the small datasets, we vary the number of layers starting from $2$ up to $15$ in the set $\{2,3,5,7,10,15\}$. For the very large datasets, we vary the number of layers up to $50$ in the set $\{2,3,5,7,10,15,20,30,40,50\}$. We also consider network widths of 16, 32, 64, or 128. %However, 
We determine the width for a dataset %as 128 
and use it for all other cases. The threshold to consider a node in a community was also varied in the set $\{0.05, 0.10, 0.125, 0.20, 0.30, 0.35, 0.40, 0.50\}$. We find the best model depth and threshold for each dataset on each of our considered methods through grid search.   Then, we run $50$ iterations with different initialization for the best hyper-parameter set. Finally, we report the mean value of the respective metric over these $50$ iterations. In the topics network, we set the number of communities to $100$ which is determined based on searching in the set $\{25, 50, 100, 200 \}$ and considering the best performing one. For the datasets having ground truth, we set the number of communities as it is in the ground truth (similar to NOCD). We also perform a statistical test to determine the statistical significance in the case of the datasets having ground truth. The following text elaborates on our baselines, datasets, metrics, statistical testing procedure, and experiment setup. Our code is available at \url{https://github.com/buet-gd/Deep-DynaResGCN-community}.

\subsection{Baselines}
In the case of the topics dataset, we use SLPA \cite{xie2011slpa}, DEMON \cite{coscia2014uncovering-demon} , MNMF \cite{wang2017community-mnmf}, BigClam \cite{yang2013overlapping-bigclam}, DANMF \cite{ye2018deep-danmf}, and NOCD \cite{shchur2019overlapping} as the baselines. We use their publicly available code for NOCD \cite{shchur2019overlapping}. For other baselines, we use the implementations in the CDlib python library with default parameters \cite{cdlib}. In the case of other datasets with ground truth, we compare our methods with the same baselines used in the NOCD \cite{shchur2019overlapping}. Our method DynaResGCN-G denotes the DynaResGCN method without considering node attributes, while DynaResGCN-X denotes the same method %with
considering node attributes. As we consider exactly the same baselines on the same datasets as described in NOCD \cite{shchur2019overlapping}, we report the results already published in \cite{shchur2019overlapping}. These baselines include BigClam \cite{yang2013overlapping-bigclam}, EPM \cite{zhou2015infinite-epm}, SNetOC \cite{todeschini2016exchangeable-snetoc}, CESNA \cite{yang2013community-cesna}, SNMF \cite{wang2011community-snmf}, CDE \cite{li2018community-cde}, DeepWalk \cite{perozzi2014deepwalk} with Non-exhaustive Overlapping K-means \cite{whang2015non-neo}(DW/NEO) , and Graph2Gauss \cite{bojchevski2017deep-g2g} with Non-exhaustive Overlapping K-means \cite{whang2015non-neo} (G2G/NEO).

% In the topic network \cite{burd2018gram}, we fixed the number of communities to 100. The value was determined over different iterations varying the number of communities and comparing the quality measure. Table \ref{tab:my-methods} shows our different methods with specific variation. Since Facebook dataset has smaller graphs, we used small community size ($\approx 20$) determined heuristically.
\subsection{Datasets}
\begin{figure}[t]
    \centering
    \includegraphics[height=0.5\textwidth,width=0.5\textwidth]{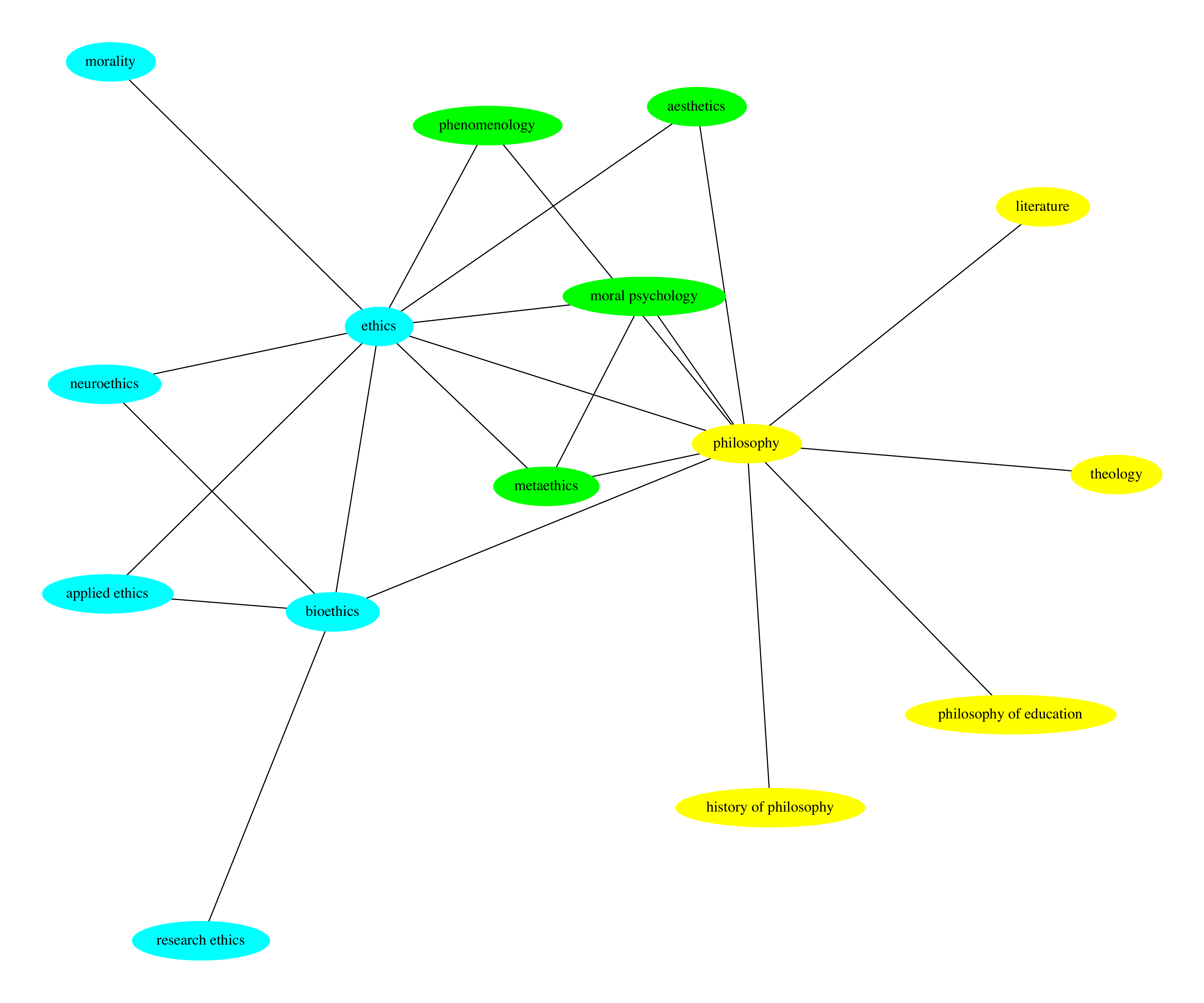}
    \caption{Two overlapping communities in a small portion of the topics network. %Green nodes belong to both communities. These green nodes are the overlapped nodes
    }
    \label{fig:small-topic}
\end{figure}

All of our datasets are described in Table \ref{tab:datasets}.
\begin{table}[h]
\caption{Our considered datasets}
\centering
\resizebox{\linewidth}{!}{%
\begin{tabular}{|l|l|l|l|l|l|l|} 
\hline
\textbf{Dataset}                        & \textbf{Nodes} & \textbf{Edges} & \textbf{MCD$^1$}& \textbf{Connectivity$^2$} & \textbf{Ground truth} & \textbf{Comment}  \\ 
\hline
Topic &     5947           &    26695 & N/A & \cmark
            &  Not available     & Research topics network   \cite{burd2018gram}         \\ 
\hline
Facebook 348 & 227 & 6384 & 4 & \xmark &Hand-labeled &Social network \cite{mcauley2014discovering-facebook-dataset} \\
\hline
Facebook 414 & 159 & 3386 & 2 & \xmark &Hand-labeled  & Social network \cite{mcauley2014discovering-facebook-dataset}\\
\hline 
Facebook 686 & 170 & 3312 & 5 & \xmark &Hand-labeled  & Social network \cite{mcauley2014discovering-facebook-dataset} \\
\hline 
Facebook 698 & 66 & 540 & 2 & \xmark &Hand-labeled  & Social network \cite{mcauley2014discovering-facebook-dataset} \\
\hline
Facebook 1684                              &       792         &   28048  & 5 & \xmark           &      Hand-labeled      & Social network  \cite{mcauley2014discovering-facebook-dataset}     \\ 
\hline
Facebook 1912                              &       755         &     60050   & 3 & \xmark        &   Hand-labeled          & Social network \cite{mcauley2014discovering-facebook-dataset}      \\

\hline
% Chemistry & 35409 & 314716 & Co-authorship network \cite{shchur2019overlapping} \\
% \hline
Computer Science & 21957 & 193500 & INF & \cmark & Empirical$^3$  & Co-authorship network \cite{shchur2019overlapping} \\
\hline 
Engineering & 14297 & 98610 & INF & \cmark &Empirical$^3$  & Co-authorship network \cite{shchur2019overlapping} \\
\hline 
Medicine & 63282 & 1620628 & INF & \cmark &Empirical$^3$  & Co-authorship network \cite{shchur2019overlapping}\\
\hline

\end{tabular}
}
\begin{tablenotes}
{\footnotesize
    \item  (1) MCD stands for the maximum community diameter. N/A refers to that ground truth is unavailable. INF refers that every community-induced sub-graph in the corresponding dataset is a disconnected graph. Therefore, MCD is undefined and theoretically infinite. %for all of the community subgraphs.
    \item (2) Connectivity says whether the whole graph is connected. %In the last 3 datasets with empirical ground truth, we observe that each of these datasets is a connected graph. However, none these datasets contains a single community induced sub-graph which is connected. This implies that the communities in these datasets contains nodes which are never directly connected to the other nodes of the same community. Thus they have very unreliable ground truth information. Indeed, these 3 datasetss are good to test scalability limits instead of detection quality.
    \item (3) %Communities are assigned based on the respective research areas.
    Community assignment is not hand-labeled by a human agent. 
    %(1) Communities are assigned based on the respective research areas. Community assignment is not hand-labeled by human agent.  %Communities correspond to research areas in respective fields,

}
\end{tablenotes}
\label{tab:datasets}
\end{table}
We use a topics graph dataset \cite{burd2018gram} that is based on various research topics. We can observe a small portion of the topics network in Figure \ref{fig:small-topic}. Every topic is a node in the graph. An edge connects two topics if a researcher works on both topics. One of the main motivations for using the topics network is using visualization to understand overlapping community structures. Most works in the literature evaluated the results of experiments on community detection by just observing the values of the evaluation metrics. In fact, the similarity/dissimilarity between two different topics is evident from the names of the topics. For instance, if we consider \emph{computer science} and \emph{algorithms}, then it is clearly evident that they should belong to the same community. On the contrary, if we consider \emph{bio-chemistry} and \emph{graph theory}, then it is also obvious that they should not belong to the same community. Thus, we can generate a heatmap of similar topics group and dissimilar topics group, which would help us understand the results %empirically/
visually. This heatmap is explained later.
% \begin{figure}[t]
%     \centering
%     \includegraphics[height=0.5\textwidth,width=0.5\textwidth]{images/small.eps}
%     \caption{Two overlapping communities in a small portion of the topics network. %Green nodes belong to both communities. These green nodes are the overlapped nodes
%     }
%     \label{fig:small-topic}
% \end{figure}
We also consider very large co-authorship networks compiled by authors of NOCD from Microsoft Academic Graph \cite{shchur2019overlapping}. The communities are assigned roughly in these datasets based on the respective research areas. In this case, hand-labeled ground truth is not available. Additionally, every community-induced subgraph is a disconnected graph, while the entire dataset is a connected graph. Thus, these datasets are %not
highly unreliable. However, these datasets may help understand the different methods' scalability limits. % in the case of very large graphs.

Finally, we consider real-world social network datasets \cite{mcauley2014discovering-facebook-dataset}. These are small datasets with reliable ground truth where communities are hand-labeled. Using these datasets, we can estimate community recovery with much better reliability. %than that of with the very large datasets without reliable ground truth. 

In essence, we experiment on datasets having different sizes %/scales 
with the number of nodes ranging from $66$ up to $65K$. Along with very different-sized datasets, we use datasets with hand-labeled ground truth, datasets with empirically assigned ground truth, and datasets without ground truth, which ensure the evaluation procedure's robustness, reliability, and strength. 

\subsection{Evaluation Metric and Visualization} In the case of the topics dataset, we consider different quality measures or fitness functions to evaluate the detected communities. These quality measures include conductance \cite{yang2013overlapping-bigclam}, clustering coefficient, density, and coverage. 
% Among these fitness functions, conductance is considered as the main evaluation metric \cite{leskovec2010empirical}. However, conductance can show very good performance in some  corner cases where the actual performance is not good at all. To detect these degenerate cases of conductance, we incorporate other metrics such as the clustering coefficient, density, and coverage. It is impossible that  corner cases with very poor actual performance would show good performance in terms of conductance and %simultaneously with 
% all other metrics.
In the case of the topics network, we also use  heatmaps %of multiple communities 
to understand the overlapping tendencies of various communities. In the case of other datasets, we use normalized mutual information (NMI) \cite{mcdaid2011normalized-nmi} to measure the similarity between ground-truth communities and predicted communities. In the case of the Facebook datasets, we visualize the largest detected community to understand the quality of the detection. To understand the definitions below, let us assume $S$ is the set of nodes in a single community. $S_1,S_2,\ldots,S_k$ denotes the set of nodes in different communities. $F$ is the affiliation matrix defined earlier. In the following text, we describe each of the measures and also the visualization processes.
\par
\textbf{Conductance:} We consider the average conductance of the detected communities (weighted by community size); The lower the conductance, the better the performance. In fact, conductance covers the intuitive definition of community \cite{yang2013overlapping-bigclam}.

\begin{align*}
\textrm{Outside}(S) &= \sum_{u \in S, v \notin S} A_{uv}\\
\textrm{Inside}(S) &= \sum_{u \in S, v \in S, v \ne u} A_{uv}\\
\textrm{Conductance}(S) &= \frac{\textrm{Outside}(S)}{\textrm{Inside}(S) + \textrm{Outside}(S)}\\
\textrm{AvgConductance}(S_1, ..., S_k) &= \frac{1}{\sum_i |S_i|}\sum_i \textrm{Conductance}(S_i) \cdot |S_i|
\end{align*}
\par
\textbf{Coverage:}
Coverage refers to the percentage of edges explained by at least one community i.e. if $(u, v)$ is an edge and both nodes share at least one community, then this edge is explained by at least one community; the higher the density, the better the performance.
\begin{align*}
   \textrm{Coverage}(S_1, ..., S_k) &= \frac{1}{|E|}\sum_{u, v \in E} \mathbf{1}[F_u^T F_v > 0] 
\end{align*}

\par
\textbf{Density:}
  The average density of the detected communities (weighted by community size); the higher the density, the better the performance.
\begin{align*}
    \rho(S) &= \frac{\text{number of existing edges in $S$}}{\text{number of possible edges in $S$}}\\
    \textrm{AvgDensity}(S_1, ..., S_k) &= \frac{1}{\sum_i |S_i|}\sum_i \rho(S_i) \cdot |S_i|
\end{align*}

\par
\textbf{Clustering Coefficient:}
The average clustering coefficient of the detected communities (weighted by community size); a higher clustering coefficient is better.
\begin{align*}
\textrm{ClustCoef}(S) &= \frac{\text{number of existing triangles in $S$}}{\text{number of possible triangles in $S$}}\\
\textrm{AvgClustCoef}(S_1, ..., S_k) &= \frac{1}{\sum_i |S_i|}\sum_i \textrm{ClustCoef}(S_i) \cdot |S_i|
\end{align*}
\par
\textbf{Normalized Mutual Information (NMI):} It is an information similarity measure between the ground truth communities and  the detected communities \cite{mcdaid2011normalized-nmi}. We use this measure when we have ground truth information.

\par
\textbf{Heatmap Visualization:}
In the case of the topics dataset, we know the names of the topics that are similar and the names of the topics that are dissimilar. To generate each heatmap, we consider a total of %9
nine topics. Among these %9
nine topics, every %3
three topics are
similar. Thus, we have %3
three groups of similar topics. However, two topics from two different groups should be dissimilar. Then, for each topic, we find the community with the strongest affiliation (as described earlier, we generate community strength for each community for each node). Now, we know the strongest community for each topic. Therefore, we have %9
nine communities, corresponding to the %9
nine topics. Then, we find the number of overlapped nodes for each pair of the communities, which can be presented as an overlapping matrix where communities are indexed in the same order, both in rows and columns. Finally, we normalize this overlapping matrix. This normalized overlapping matrix is shown as a heatmap. Each cell of the heatmap shows the overlapping strength between the corresponding communities. The higher the overlapping, the darker the cell.

\subsection{Statistical Significance Test}
We use NMI in the case of the datasets having ground truth. We use a statistical $t$-test to measure the significance of the difference in mean NMI between NOCD and DynaDesGCN. We perform a standard $t$-test with $95\%$ confidence (significance $\alpha = 0.05$) to test whether the difference between two means from two different distributions is statistically significant \cite{t-test,t-test2}. Let $\Bar{x_1}$ be the mean over $n_1=50$ (as we run 50 independent initializations for each method) independent samples from one method where $s_1$ is the standard error and $\Bar{x_2}$ is the mean over $n_2=50$ different samples from another method where $s_2$ is the standard error. Then, our null hypothesis ($H_0$) is that the difference in the mean is zero ($\mu_1 = \mu_2$) or statistically insignificant. The alternative hypothesis is that the difference in the mean is statistically %different
significant ($\mu_1 \not = \mu_2$). %or $\mu_1 > \mu_2 $ when we assume $\Bar{x_1} > \Bar{x_2}$. 
Now, we calculate the standard $t$-statistic as follows:
\begin{align*}
    t &= \frac{\Bar{x_1}- \Bar{x_2} }{\sqrt{\frac{s_1^2 }{n_1} + \frac{s_2^2}{n_2}}}
\end{align*}
For a $t$-test with a significance level of $\alpha=0.05$, we calculate upper sided %(as we assume $x_1>x_2$) 
critical value from the $t$-distribution with degrees of freedom $49$ ($\min(n_1-1, n_2-1)$). % (as we run 50 different independent initializations). 
Let us assume $t^*$ is the $\alpha/2$ upper critical value ($(1-\alpha/2)*100=97.5$th percentile). If the absolute $t$-statistic is greater than the $t^*$, then we reject the null hypothesis and state that the difference of means is statistically significant.

\subsection{Experiment Setup}
We perform our experimentation on NVIDIA GeForce RTX GPU in the Linux platform with  4GB RAM. We have implemented our codes in the PyTorch library with Python version 8. We use weight decay rate as $0.01$, learning rate $0.001$, batch-normalization, and stochastic loss \cite{shchur2019overlapping} based on the sampled edges of the original graph during training. 
% Hidden layer width in $128, 64, 32$ and network depth upto $50$ layers. 
We also use the early stopping criterion for training with a maximum $500$ epochs. %We will publish our code after the acceptance of the paper.

\section{Results and Discussion} 
A previous study \cite{shchur2019overlapping} demonstrated the effectiveness of graph neural networks for detecting overlapping communities. They showed that GNNs perform better than free variable optimization or other direct methods such as BigClam \cite{shchur2019overlapping,yang2013overlapping-bigclam}, EPM \cite{zhou2015infinite-epm}, SNetOC, \cite{todeschini2016exchangeable-snetoc} etc. In this study, we aim to develop a better GNN-based method to detect overlapping communities with higher detection quality. Specifically, we design a deep GCN encoder considering dynamic dilated aggregation for the overlapping community detection model. % \cite{shchur2019overlapping}. 
Eventually, we achieve significantly better performance and detection quality than related studies. For evaluation, we use three different types of datasets. In the following text, we describe our results. %for each of the type:
\subsection{Dataset without Ground Truth}
Table \ref{table:topic} compares our various methods' performance with the existing methods on topics dataset \cite{burd2018gram} which has no ground truth. In practice,
\begin{table}[h]
\caption{\textbf{Results from Topic dataset}}
\label{table:topic}
\centering
\resizebox{\linewidth}{!}{%
\begin{tabular}{lllll} 
\hline
Method                                                  & Conductance   ($\downarrow$)                          & clustering coefficient ($\uparrow$)          & Density ($\uparrow$)                               & Coverage    ($\uparrow$)                            \\ 
\hline

% {\cellcolor[rgb]{0.914,0.886,0.886}}DynaResGCN+cond              & {\cellcolor[rgb]{0.953,0.929,0.929}}0.2144 & {\cellcolor[rgb]{0.929,0.91,0.91}}0.00021 & {\cellcolor[rgb]{0.941,0.918,0.918}}0.00759 & {\cellcolor[rgb]{0.953,0.925,0.925}}0.9984  \\ 

\rowcolor[rgb]{0.922,0.898,0.898} DynaResGCN                     & \textbf{0.1807}                            & 0.00037                                   & 0.00781                                     & 0.9986                                      \\ 

% \rowcolor[rgb]{0.922,0.898,0.898} DynaResGCN+weight              & 0.1957                                     & 0.00023                                   & 0.00767                                     & 0.9989                                      \\ 

% \rowcolor[rgb]{0.933,0.906,0.906} DynaResGCN+differential weight & 0.2126                                    & 0.00019                                   & 0.00652                                     & 0.9985                                     \\ 

NOCD \cite{shchur2019overlapping}                                                            & 0.3049                                    & 0.00016                                   & 0.01135                                    & \textbf{0.9999}                             \\ 

SLPA      \cite{xie2011slpa}                                                       & 0.3058                                     & \textbf{0.00217}                           & \textbf{0.04867}                            & 0.8273                                        \\ 

BigClam        \cite{yang2013overlapping-bigclam}                                                  & 0.2129                                      & 0.00000                                      & 0.00175                                     & 0.8698                                      \\ 

DANMF     \cite{ye2018deep-danmf}                                                       & 0.2753                                     & 0.00008                                      & 0.00869                                     & 0.7250                                       \\ 

DEMON                   \cite{coscia2014uncovering-demon}                                         & 0.4873                                       & 0.00151                                   & 0.04199                                     & 0.8965                                      \\ 

MNMF \cite{wang2017community-mnmf}                                                             & 0.3965                                     & 0.00009                                      & 0.00907                                        & 0.6179                                      \\
\bottomrule
\end{tabular}
}
\end{table}
there %are
is no sufficiently large dataset with reliable overlapping ground truth information. Overlapping ground truth information is also more challenging because of subjectivity. It is not impossible to have different ground truths for the same dataset if the dataset is labeled based on two different perspectives. Moreover, the amount of overlap between two communities is also subjective. %Indeed, large graphs with hand-labeled ground truth are not openly available. %Therefore, though some authors \cite{shchur2019overlapping} claim they have overlapping ground truth datasets with large graphs, these are not reliable. 
%Moreover, these datasets are also not hand-labeled. %In fact, we are unable to reproduce the results on large datasets with overlapping ground truth information provided by \cite{shchur2019overlapping}. 
As a result, it is necessary to evaluate different methods on %unsupervised (having no ground truth) 
datasets having no ground truth with reliable evaluation metrics. %In fact, 
It is also good to assess with some auxiliary metrics %or fitness functions 
to detect the corner cases. Therefore, %in this study, 
we use conductance as the main evaluation metric \cite{yang2013overlapping-bigclam}, as well as three auxiliary metrics to determine whether the main metric (conductance) is actually performing well. Finally, we attempt to visualize overlapping tendencies using a heatmap.
\begin{figure}[htp]
     \centering
     
          \begin{subfigure}[b]{0.4\textwidth}
         \centering
         \includegraphics[width=\textwidth]{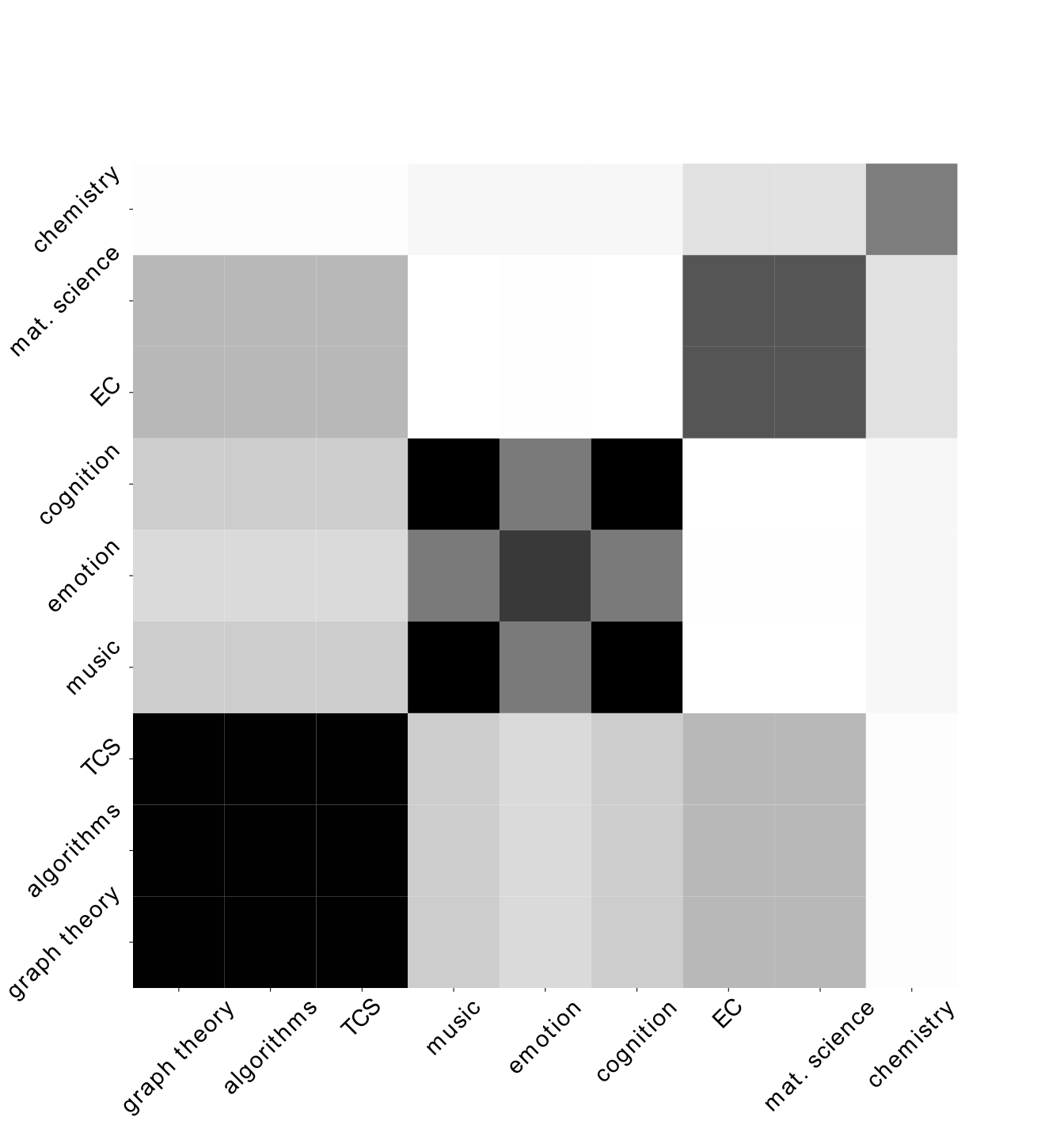}
         \caption{DynaResGCN (\textbf{ours})}
         \label{fig:deep-unweighted-heatmap}
     \end{subfigure}
     \begin{subfigure}[b]{0.4\textwidth}
         \centering
         \includegraphics[width=\textwidth]{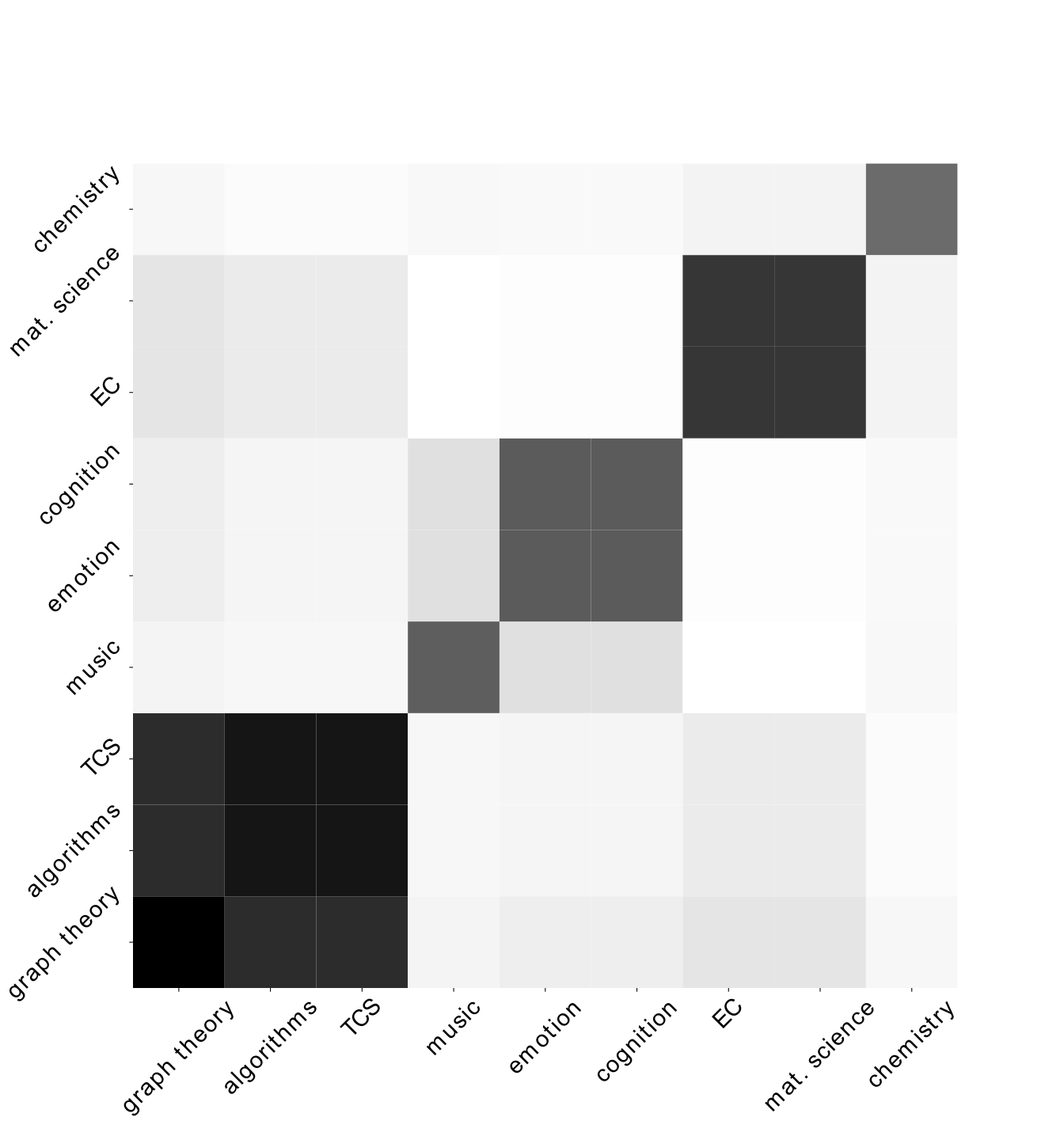}
          \caption{NOCD }
         \label{fig:shallow-heatmap}
     \end{subfigure}
    %  \hfill
      \begin{subfigure}[b]{0.4\textwidth}
         \centering
         \includegraphics[width=\textwidth]{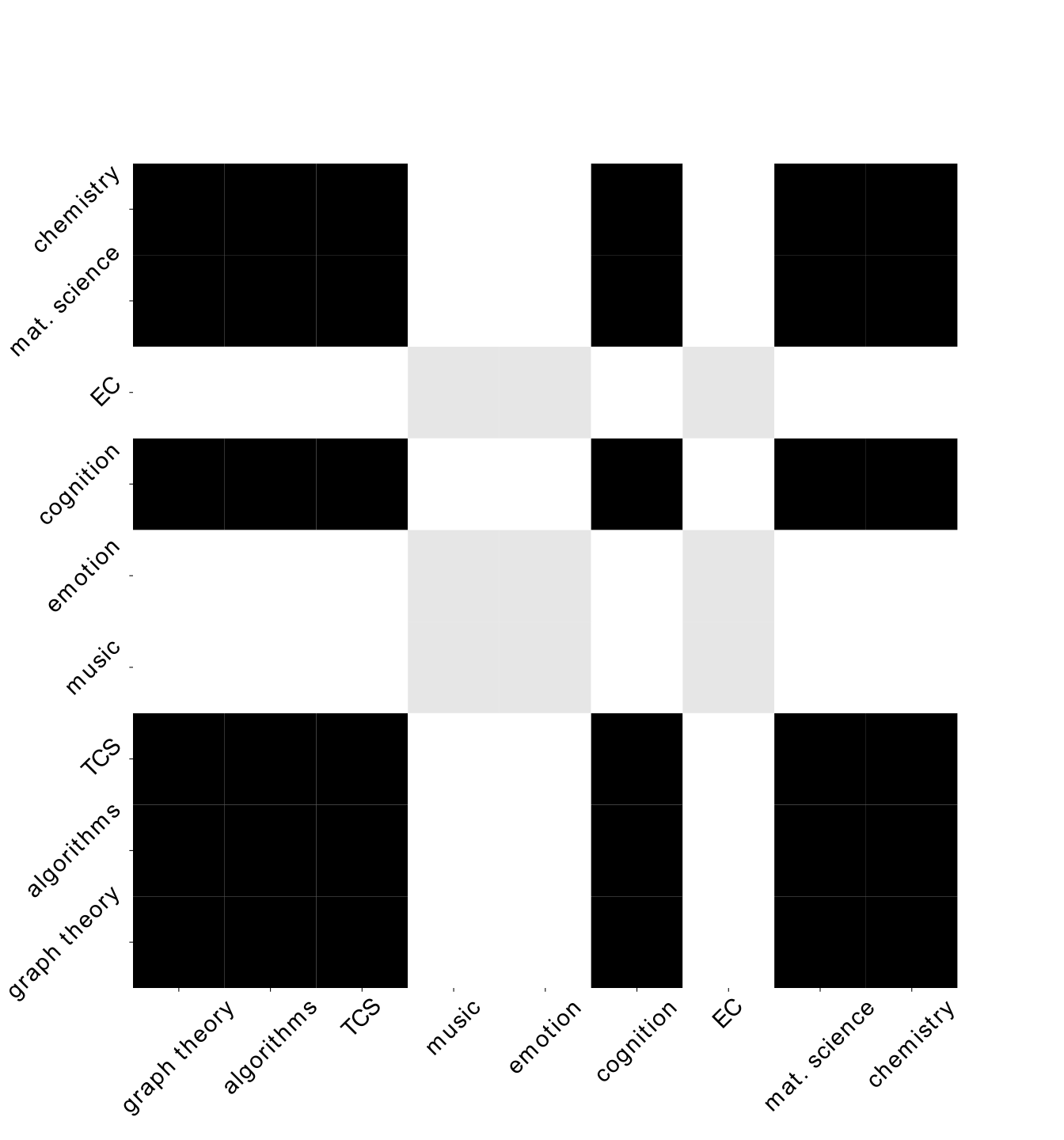}
         \caption{BigClam}
         \label{fig:bigclam-heatmap}
     \end{subfigure}
    \begin{subfigure}[b]{0.4\textwidth}
         \centering
         \includegraphics[width=\textwidth]{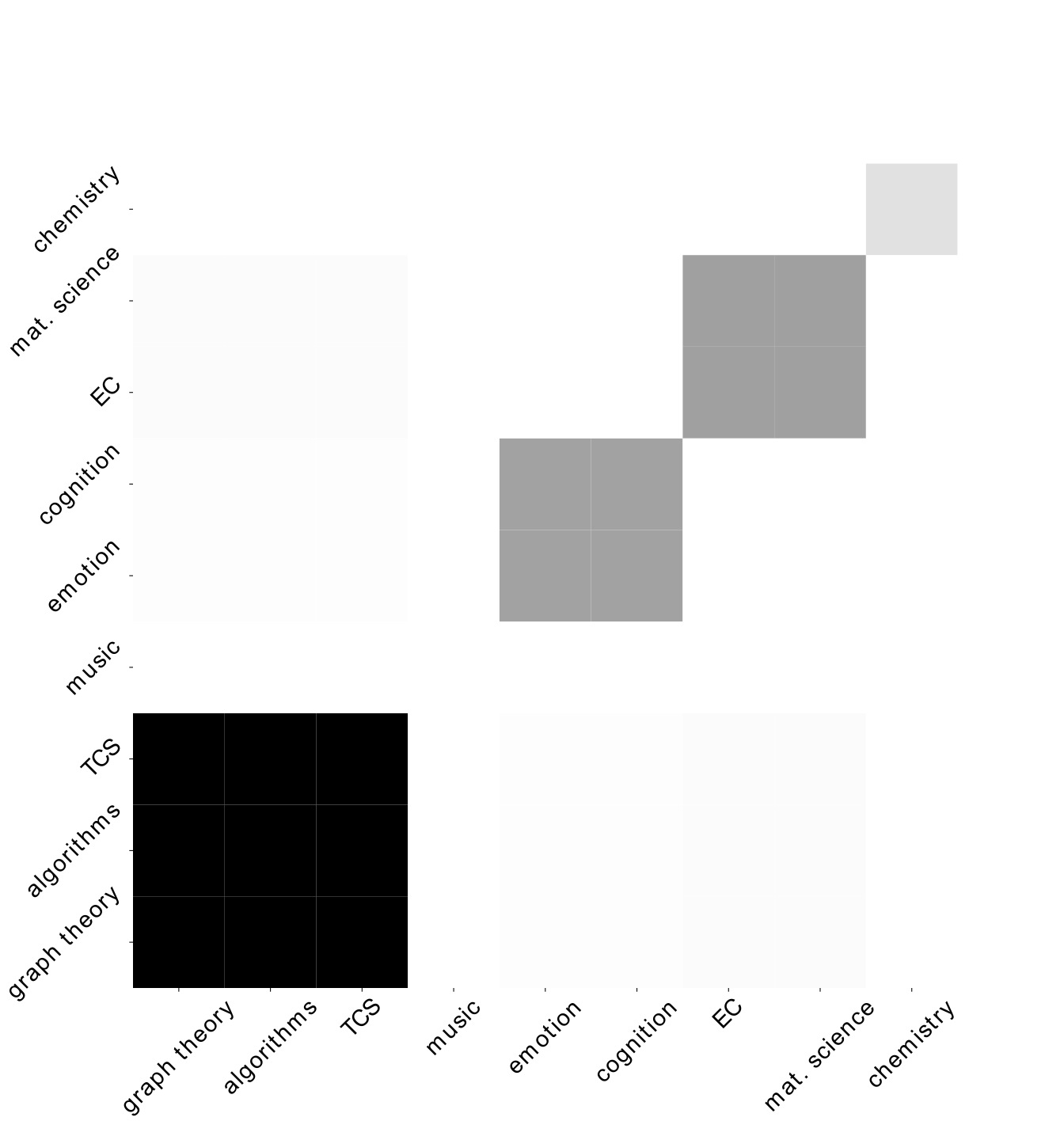}
         \caption{SLPA}
         \label{fig:SLPA-heatmap}
     \end{subfigure}
    \begin{subfigure}[b]{0.4\textwidth}
         \centering
         \includegraphics[width=\textwidth]{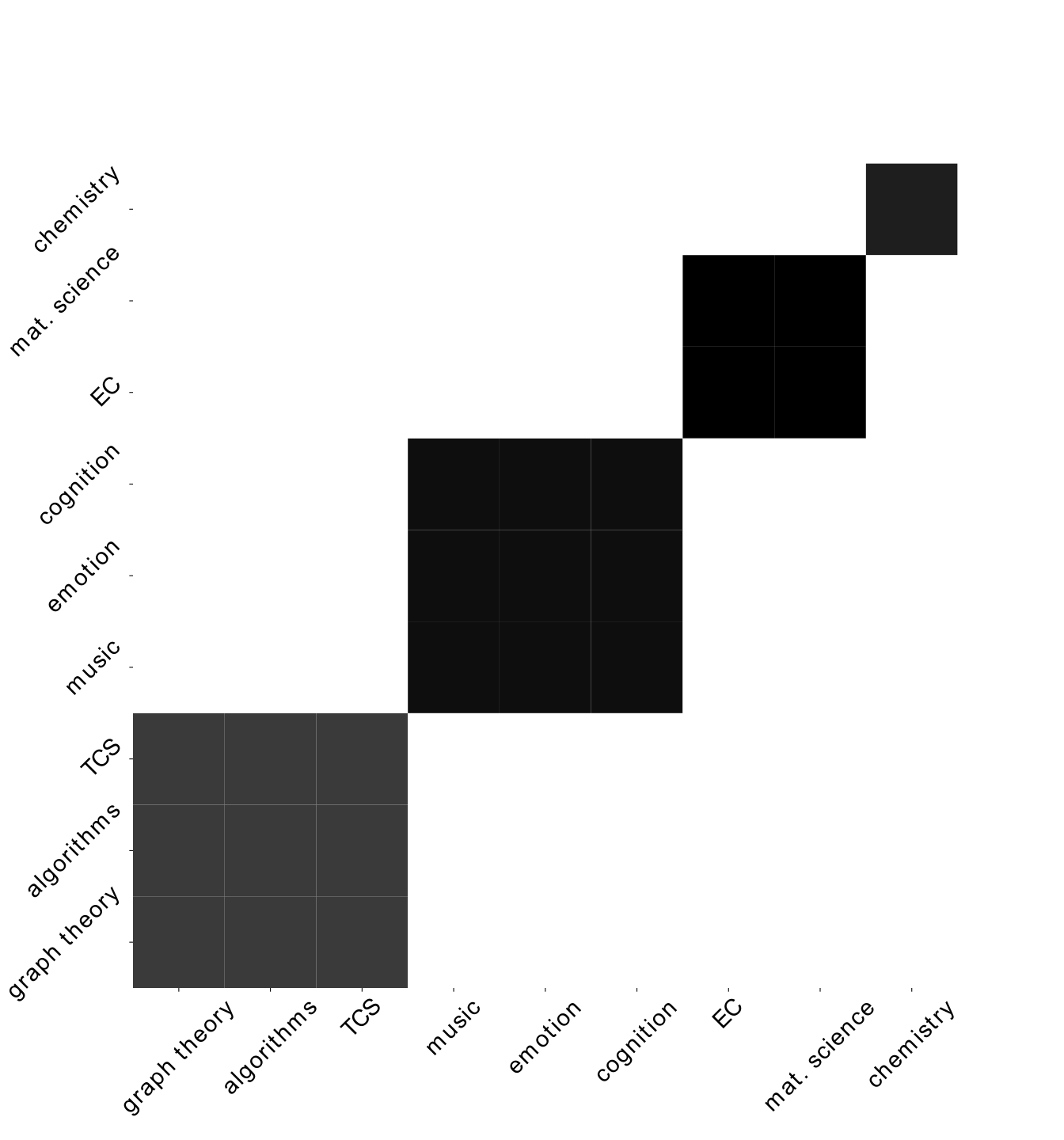}
         \caption{NMNF}
         \label{fig:NMNF-heatmap}
     \end{subfigure}
     \begin{subfigure}[b]{0.4\textwidth}
         \centering
         \includegraphics[width=\textwidth]{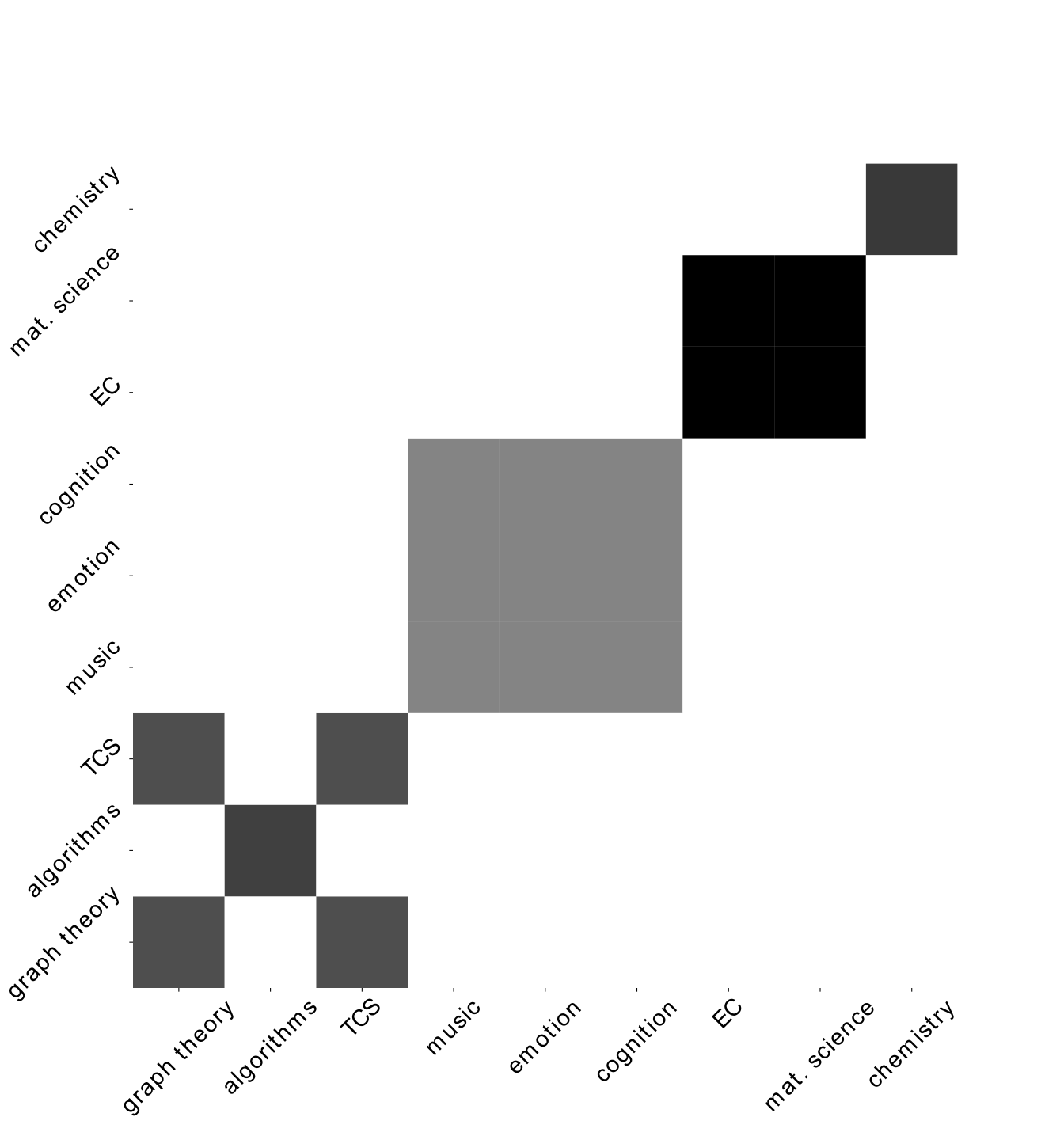}
         \caption{DANMF}
         \label{fig:DANMF-heatmap}
     \end{subfigure}
  
        \caption{Overlapping tendency of  selected topics from the topics dataset. %based on different methods. 
        The darker the cell, the higher the overlap.}
        \label{fig:heatmap}
\end{figure}

We compare our method \emph{DynaResGCN} %, \emph{DynaResGCN+cond}, \emph{DynaResGCN+weight}, and \emph{DyanResGCN+differential weight}
with NOCD \cite{shchur2019overlapping}, SLPA \cite{xie2011slpa}, BigClam \cite{yang2013overlapping-bigclam}, DANMF \cite{ye2018deep-danmf}, DEMON \cite{coscia2014uncovering-demon}, and MNMF \cite{wang2017community-mnmf} on the topics dataset (Table \ref{table:topic}). %The results are presented in Table \ref{table:topic}. 
Our %best 
method DynaResGCN performs the best in terms of conductance and also comparatively well in terms of %the 
other auxiliary metrics.
%In terms of conductance, DynaResGCN  performs the best (the lower the conductance, the better the performance).
Indeed, conductance is the most important metric because it captures the basic definition of a community \cite{yang2013overlapping-bigclam}.
Bigclam has the closest performance to DynaResGCN in terms of conductance. However, in terms of clustering coefficient and density, BigClam performs the worst. The performance of BigClam is also not good in terms of coverage. Therefore, %the performance of BigClam in this case 
it is a degenerate case. %Among all of the metrics, we consider

Instead of relying on just numbers, we attempt to understand overlapping community detection through a %novel 
heatmap visualization approach. We choose three groups of topics to create heatmaps based on the topic network dataset. The first group consists of \emph{graph theory}, \emph{algorithms}, and \emph{theoretical computer science (TCS)}. The second group consists of \emph{music}, \emph{emotion}, and \emph{cognition}. The third group consists of \emph{electro-chemistry (EC)}, \emph{material science (mat. science)}, and \emph{chemistry}. In the heatmap in Figure \ref{fig:heatmap}, the topics are arranged in order of the first group, second group, and third group. % with index starting from $0$ to $9$. 

The heatmaps show the effectiveness of each method for detecting overlapping regions. It is %evident
visible from the heatmaps that the GNN-based models are very effective for detecting overlapping communities. It is also evident from the heatmaps that the DynaResGCN-based models better detect overlapping communities than the shallow models. There is an interesting observation related to the BigClam model. Based on conductance, BigClam performed very well. However, in terms of the other metrics, BigClam performed poorly. The heatmaps clearly reveal that the performance of BigClam based on conductance is a corner case or degenerate case, where conductance does not reflect the true phenomenon. Therefore, based on the heatmaps, we confidently conclude that if a method performs well in terms of conductance and also in all other metrics, \emph{only} then we can refer that the method is a %good/better
winning method. 
\par 
\subsection{Datasets with Ground Truth}
We compare our best method, the DynaResGCN model, on a set of hand-labeled %graphs/
networks and on a set of very large graphs with empirically assigned ground truth. In both of the cases, we compare %the 
DynaResGCN and NOCD with attributes of the nodes and without attributes of the nodes in terms of NMI. To evaluate in terms of NMI, we require ground truth. The methods DynaResGCN-G and NOCD-G are compared without considering the attributes of the nodes in Table \ref{tab:ground-truth-G}, and the details on DynaResGCN-G are described in Table \ref{tab:dyna-G-details}. On the other hand, the methods DynaResGCN-X and NOCD-X are compared considering node attributes in Table \ref{tab:ground-truth-x} and details on DynaResGCN-X are described in Table \ref{tab:dyna-x-details}. We explain the results in the following text:
\begin{table}[hb]
\caption{Results for community recovery in terms of overlapping NMI without node attributes using 50 different initializations. %Moreover, we decide the significance of the mean difference of NMI between NOCD-G and DynaResGCN-G with $95\%$ confidence based on the standard $t$-test.}
}
\centering
\resizebox{\linewidth}{!}{%
\begin{tabular}{llllllllllll}
\hline
Dataset          & BigCLAM & CESNA & EPM  & SNetOC & CDE  & SNMF & DW/NEO & G2G/NEO & NOCD-G & DynaResGCN-G & Significant$^1$                    \\ 
\hline
Facebook 348     & 26.0    & 29.4  & 6.5  & 24.0   & 24.8 & 13.5 & 31.2   & 17.2    & 34.7   & \textbf{39.8 } & \cmark  \\
Facebook 414     & 48.3    & 50.3  & 17.5 & 52.0   & 28.7 & 32.5 & 40.9   & 32.3    & 56.3   & \textbf{58.1 }  & \cmark  \\
Facebook 686     & 13.8    & 13.3  & 3.1  & 10.6   & 13.5 & 11.6 & 11.8   & 5.6     & 20.6   & \textbf{25.4  } & \cmark \\
Facebook 698     & 45.6    & 39.4  & 9.2  & 44.9   & 31.6 & 28.0 & 40.1   & 2.6     & 49.3   & \textbf{51.0} & \cmark  \\
Facebook 1684    & 32.7    & 28.0  & 6.8  & 26.1   & 28.8 & 13.0 & 37.2   & 9.9     & 34.7   & \textbf{44.3  } & \cmark  \\
Facebook 1912    & 21.4    & 21.2  & 9.8  & 21.4   & 15.5 & 23.4 & 20.8   & 16.0    & 36.8   & \textbf{40.1 }  & \cmark  \\
% Chemistry        & 0.0     & 23.3  & DNF  & DNF    & DNF  & 2.6  & 1.7    & 22.8    & 22.6   &                                  \\
Computer Science & 0.0     & 33.8  & DNF  & DNF    & DNF  & 9.4  & 3.2    & 31.2    & 34.2   &   \textbf{37.4} & \cmark                               \\
Engineering      & 7.9     & 24.3  & DNF  & DNF    & DNF  & 10.1 & 4.7    & 33.4    & 18.4   &         \textbf{37.3}  & \cmark                      \\
Medicine         & 0.0     & 14.4  & DNF  & DNF    & DNF  & 4.9  & 5.5    & 28.8    & 27.4   &    \textbf{37.3}    & \cmark                       \\
\bottomrule
\end{tabular}
}
\label{tab:ground-truth-G}
\begin{tablenotes}
{\footnotesize
    \item  (1) The significance of the mean difference of NMI between NOCD-G and DynaResGCN-G  is determined with $95\%$ confidence based on the standard $t$-test.

}
\end{tablenotes}
\end{table}

\begin{table}
\centering
\caption{DynaResGCN-G experimental details and results}
\resizebox{0.55\linewidth}{!}{%
\begin{tabular}{lllllll}
\hline
Dataset           & Mean NMI & Std error (NMI) & Layers & Threshold  \\
\hline
Facebook 348           & 39.80    & $\pm2.2$             & 7      & 0.40       \\
Facebook 414          & 58.02    & $\pm3.2$             & 2      & 0.50       \\
Facebook 686           & 25.40    & $\pm1.9$             & 7      & 0.35       \\
Facebook 698          & 51.00    & $\pm3.3$             & 5      & 0.50       \\
Facebook 1684         & 44.30    & $\pm2.6$             & 3      & 0.50       \\
Facebook 1912          & 40.10    & $\pm1.7$             & 3      & 0.50       \\
% Chemistry               &          &                 &        &            \\
Computer Science        &       37.4   &    $\pm{1.7}$             &    15    & 0.50            \\
Engineering             &    37.3     &       $\pm{1.2}$          &   15     &       0.35     \\
Medicine                &  37.3        &    $\pm{1.1}$             &  15       &    0.35       \\
\bottomrule
\end{tabular}
}
\label{tab:dyna-G-details}

\end{table}

\begin{table}
\centering
\caption{Results for community recovery in terms of overlapping NMI with node attributes using 50 different initializations. %Moreover, we decide the significance of the mean-difference of NMI between NOCD-X and DynaResGCN-X with $95\%$ confidence based on the standard $t$-test.
}
\resizebox{\linewidth}{!}{%
\begin{tabular}{llllllllllll}
\hline
Dataset                    & BigCLAM & CESNA & EPM  & SNetOC & CDE  & SNMF & DW/NEO & G2G/NEO & NOCD-X & DynaResGCN-X & Significant$^1$ \\ 
\hline
Facebook 348               & 26.0    & 29.4  & 6.5  & 24.0   & 24.8 & 13.5 & 31.2   & 17.2    & 36.4   &  \textbf{39.8}           & \cmark         \\
Facebook 414               & 48.3    & 50.3  & 17.5 & 52.0   & 28.7 & 32.5 & 40.9   & 32.3    & \textbf{59.8}   & 
\textbf{59.0} & \xmark \\
Facebook 686               & 13.8    & 13.3  & 3.1  & 10.6   & 13.5 & 11.6 & 11.8   & 5.6     & 21.0   & \textbf{27.3}       & \cmark      \\
Facebook 698               & 45.6    & 39.4  & 9.2  & 44.9   & 31.6 & 28.0 & 40.1   & 2.6     & 41.7   & \textbf{50.1}        & \cmark      \\
Facebook 1684              & 32.7    & 28.0  & 6.8  & 26.1   & 28.8 & 13.0 & 37.2   & 9.9     & 26.1   & \textbf{41.0}     & \cmark        \\
Facebook 1912              & 21.4    & 21.2  & 9.8  & 21.4   & 15.5 & 23.4 & 20.8   & 16.0    & 35.6   & \textbf{39.6}   & \cmark           \\
% Chemistry & 0.0     & 23.3  & DNF  & DNF    & DNF  & 2.6  & 1.7    & 22.8    & 45.3   &               \\
Computer Science           & 0.0     & 33.8  & DNF  & DNF    & DNF  & 9.4  & 3.2    & 31.2    & \textbf{50.2}   &    46.0 & \cmark         \\
Engineering                & 7.9     & 24.3  & DNF  & DNF    & DNF  & 10.1 & 4.7    & 33.4    & \textbf{39.1}   &    \textbf{39.0}     &  \xmark    \\
Medicine                   & 0.0     & 14.4  & DNF  & DNF    & DNF  & 4.9  & 5.5    & 28.8    & 37.8   &     \textbf{40.0}   & \cmark      \\
\bottomrule
\end{tabular}
}
\label{tab:ground-truth-x}
\begin{tablenotes}
{\footnotesize
    \item  (1) The significance of the mean difference of NMI between NOCD-G and DynaResGCN-G is determined with $95\%$ confidence based on the standard $t$-test.
}
\end{tablenotes}
\end{table}

\begin{table}
\centering
\caption{DynaResGCN-X experimental details and results}
\resizebox{0.55\linewidth}{!}{%
\begin{tabular}{lllllll}
\hline
Dataset           & Mean NMI & Std error (NMI) & Layers & Threshold  \\
\hline
Facebook 348           & 39.8    & $\pm2.2$             & 7      & 0.35       \\
Facebook 414          & 59.0    & $\pm3.1$             & 3      & 0.50       \\
Facebook 686           & 27.3    & $\pm2.0$             & 5      & 0.125       \\
Facebook 698          & 50.1    & $\pm2.3$             & 7      & 0.50       \\
Facebook 1684         & 41.0    & $\pm2.1$             & 7      & 0.50       \\
Facebook 1912          & 39.6    & $\pm1.7$             & 3      & 0.50       \\
% Chemistry               &          &                 &        &            \\
Computer Science        &     46.0     &    $\pm{2.2}$             &    2    &  0.50           \\
Engineering             &  39.0        &          $\pm{3.3}$       & 40       &   0.35         \\
Medicine                &   40.0       &  $\pm{2.1}$               & 2       &     0.35      \\
\bottomrule 
\end{tabular}
}
\label{tab:dyna-x-details}
\end{table}
\par
\textbf{Dataset with Hand-labeled Ground Truth:} 
These are small graphs from Facebook \cite{mcauley2014discovering-facebook-dataset} with reliable (hand-labeled) ground truth. From Table \ref{tab:ground-truth-G}, and Table \ref{tab:ground-truth-x}, it is clearly evident that our method DynaResGCN significantly outperforms the nearest best method NOCD in all of the Facebook networks irrespective of the node attributes. In fact, only in the case of the dataset \emph{Facebook 414}, NOCD-X has a comparative performance with DynaResGCN-X. DynaResGCN outperforms all the other methods in these datasets with ground truth by a large margin. 

%\textcolor{red}{
One of the key implications of our method is the robustness to node features. In fact, in some cases, DynaResGCN-G outperforms DynaResGCN-X. For instance, in the datasets including \textit{Facebook 348, Facebook 698, Facebook 1684, and Facebook 1912}, DynaResGCN-G is surprisingly robust, outperforming all the methods even with node attributes. This implies that our method can exploit structural information lying in the graph very well. From Table \ref{tab:dyna-G-details} and Table \ref{tab:dyna-x-details}, it is prominent that different datasets have different model depths and thresholds, which in turn proves that model depth and thresholds are subjective to the datasets.
%}

%\textcolor{red}{
Achieving state-of-the-art performance from deeper models implies that our method can train deep GCNs successfully in general irregular graphs. This proves that DynaResGCN can resolve the over-smoothing problem and vanishing gradient problem. %effectively. 
It also proves that we effectively incorporate the concepts of residual connection, dilated aggregation, and dynamic edges in our DynaResGCN model. 
%}

\begin{figure}[th]
     \centering
     \begin{subfigure}[b]{0.45\textwidth}
         \centering
         \includegraphics[width=\textwidth]{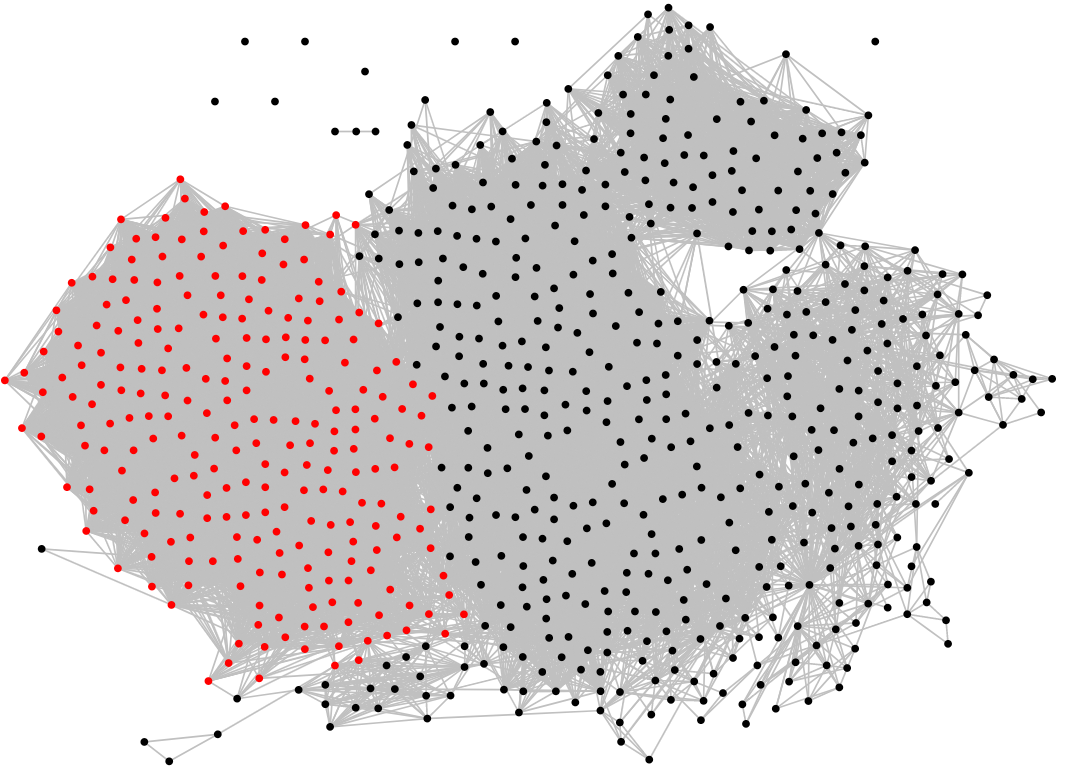}
          \caption{DynaResGCN (\textbf{ours})}
         \label{fig:deep-facebook-largest}
     \end{subfigure}
    %  \hfill
     \begin{subfigure}[b]{0.45\textwidth}
         \centering
         \includegraphics[width=\textwidth]{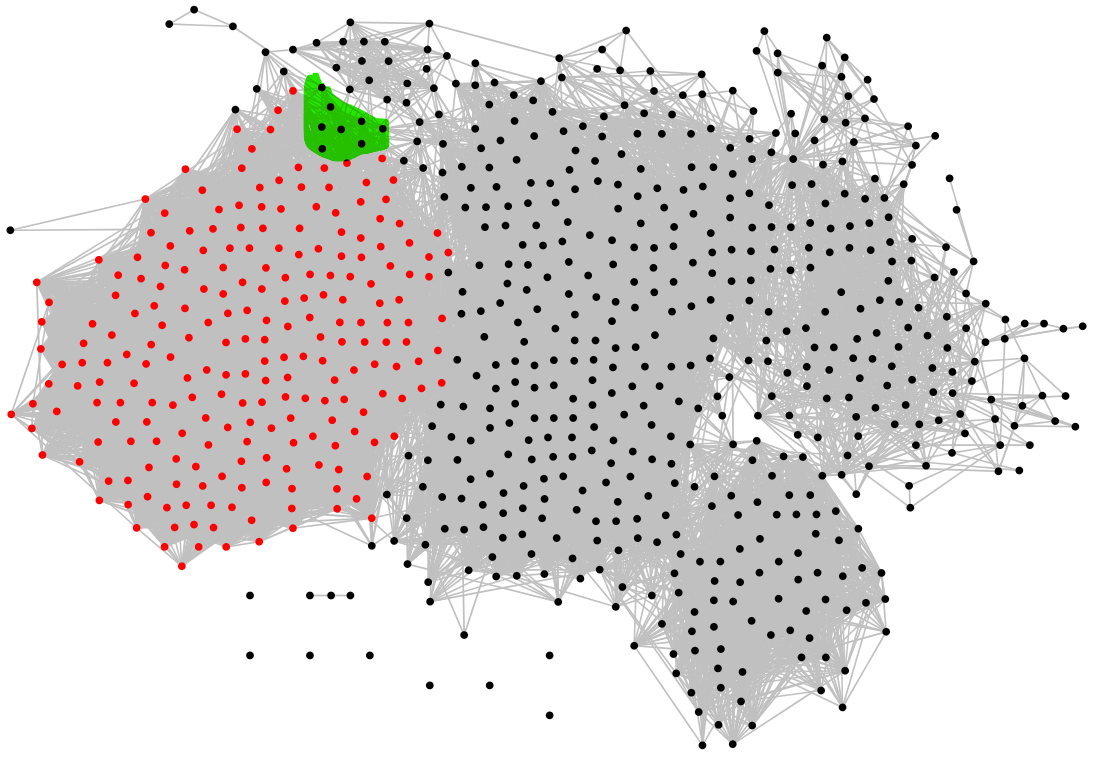}
         \caption{NOCD model}
         \label{fig:shallow-facebook-largest}
     \end{subfigure}
    %  \hfill
     \begin{subfigure}[b]{0.45\textwidth}
         \centering
         \includegraphics[width=\textwidth]{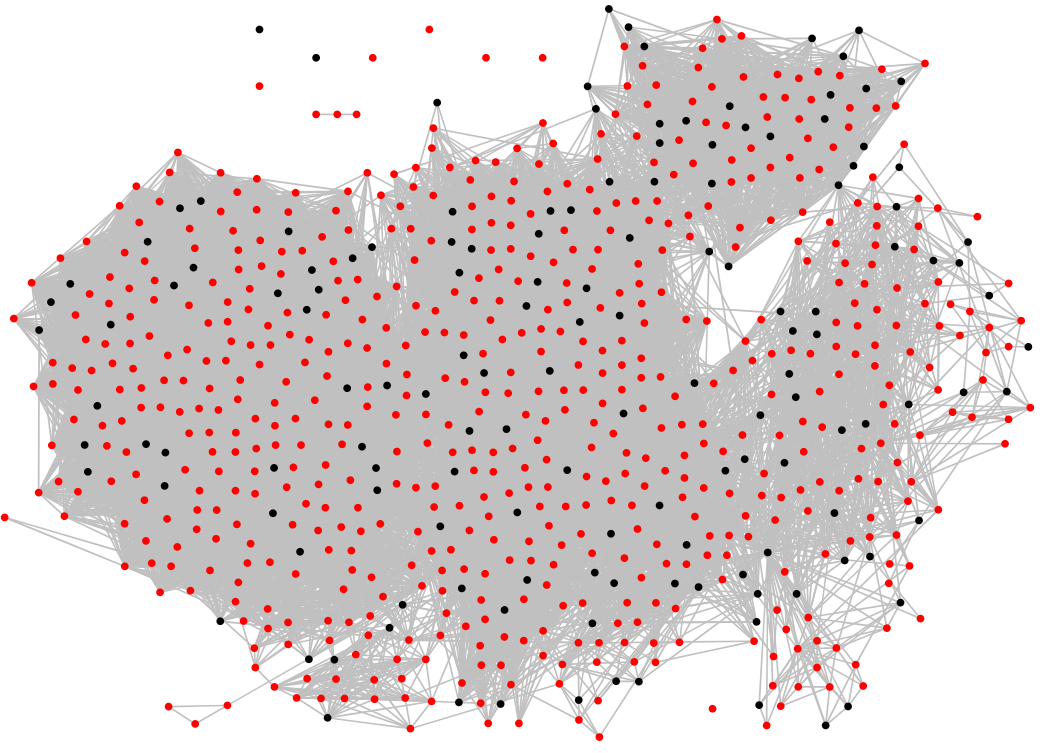}
         \caption{DEMON}
         \label{fig:DEMON-facebook-largest}
     \end{subfigure}
      \begin{subfigure}[b]{0.45\textwidth}
         \centering
         \includegraphics[width=\textwidth]{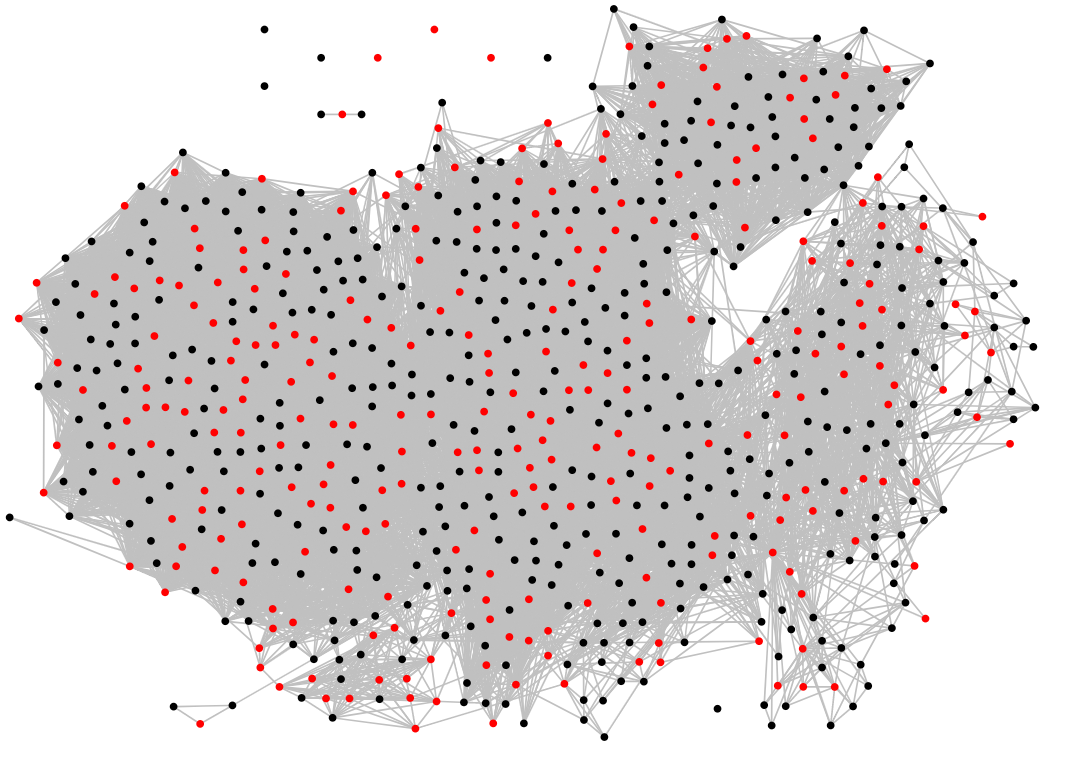}
         \caption{SLPA}
         \label{fig:SLPA-facebook-largest}
     \end{subfigure}
    
        \caption{The largest community identified by different methods in a \emph{Facebook %1912
        } dataset}
        \label{fig:point-largest-facebook}
\end{figure}

We also attempt to understand the results from the Facebook dataset %with 
by visualizing the largest cluster. Interestingly, the GNN-based methods can clearly identify the largest cluster. Among the GNN-based methods, the DynaResGCN-X model %can %identfiy 
identified the largest cluster more clearly (Figure \ref{fig:deep-facebook-largest}). The NOCD-X (GNN-based) model can identify the largest cluster in a fairly good manner. However, in Figure \ref{fig:shallow-facebook-largest}, we mark some part that should be in the largest cluster but is excluded from the largest cluster by NOCD. Other methods like SLPA and DEMON clearly fail %s 
to identify the largest cluster as in Figure \ref{fig:DEMON-facebook-largest}, \ref{fig:SLPA-facebook-largest}. It is evident that the DynaResGCN-based method is clearly the best method to identify overlapping communities in % the
%graphs/
networks.

\par
\textbf{Datasets with Empirical Ground Truth:} These datasets include \emph{Computer Science}, \emph{Engineering}, and \emph{Medicine}. These are very large datasets having nodes up to $65K$. Reliable ground truth information is not available for these datasets. As these are co-authorship networks, ground truth is roughly assigned based on the research area. In this case, DynaResGCN outperforms NOCD by a large margin when node attributes are ignored. When node attributes are considered, NOCD can outperform DynaResGCN only in a single case. In most other cases, DynaResGCN significantly outperforms NOCD. Moreover, it also proves that DynaResGCN is much more robust to node attributes. In fact, in some cases, DynaResGCN can reach the best performance even without considering node attributes. 
%\textcolor{red}{
Indeed, we experiment with these datasets to test the scalability limits of our methods. Surprisingly, our DynaResGCN-based methods are highly scalable with very deep models. Table \ref{tab:execution-time} shows that our methods can converge in less than 2 minutes in almost every dataset.

\begin{table}[th]
\centering
\caption{Comparison of the execution time of NOCD and DynaResGCN without node attributes. The time of DynaResGCN is reported for the best number of layers as in table \ref{tab:dyna-G-details}. All the times are reported in seconds.}
\resizebox{0.55\linewidth}{!}{%
\begin{tabular}{llll} 
\hline
\multirow{2}{*}{Dataset} & \multirow{2}{*}{NOCD} & \multicolumn{2}{l}{~ ~ ~ ~ ~ ~ ~ ~ ~ DynaResGCN}  \\ 
\cline{3-4}
                         &                       & ~training~ & ~ pre-processing + training          \\ 
\hline
Facebook 348             & ~ ~6                 & ~ ~12     & ~ ~ ~ ~ ~ ~ ~ ~ ~ 12                 \\
Facebook 414             & ~ ~5                 & ~ ~ 5      & ~ ~ ~ ~ ~ ~ ~ ~ ~ ~5                 \\
Facebook 686             & ~ ~5                 & ~ ~ 8~ ~   & ~ ~ ~ ~ ~ ~ ~ ~ ~ ~8                 \\
Facebook 698             & ~ ~6                  & ~ ~ 9      & ~ ~ ~ ~ ~ ~ ~ ~ ~ ~9                 \\
Facebook 1684            & ~ ~6                  & ~ ~ 8      & ~ ~ ~ ~ ~ ~ ~ ~ ~ ~8                 \\
Facebook 1912            & ~ 10                  & ~ ~12      & ~ ~ ~ ~ ~ ~ ~ ~ ~ 12                 \\
Computer Science         & ~ 13                  & ~ ~109     & ~ ~ ~ ~ ~ ~ ~ ~ ~113                 \\
Engineering              & ~ 10                  & ~ ~63      & ~ ~ ~ ~ ~ ~ ~ ~ ~ 65                 \\
Medicine                 & ~ 28                  & ~ ~759     & ~ ~ ~ ~ ~ ~ ~ ~ ~808   \\
\bottomrule 
\end{tabular}
}
\label{tab:execution-time}
\end{table}
\subsection{Computational Efficiency and Execution Time}
The execution times for NOCD and DynaResGCN are reported in Table \ref{tab:execution-time}. In this table, pre-processing time refers to the time for graph augmentation %done by algorithms \ref{alg:random-augmentation}, \ref{alg:weighted-augmentation} 
before training. In the case of the DynaResGCN method, the execution time is reported for the optimal number of layers for each dataset as it is in Table \ref{tab:dyna-G-details}. For the smaller datasets from Facebook, the execution time for DynaResGCN is less than 15 seconds, even with a deep model %with up to
having seven layers. In the case of large datasets up to $65$ thousand nodes, the execution time of DynaResGCN is less than 15 minutes with a deep model up to 15 layers. In fact, the execution time of DynaResGCN is less than 2 minutes, even with 15 layers except for the largest dataset (\emph{Medicine}). The reason is actually behind our design of deep GCN, where we consider at most two hops of the neighborhood, making our model efficient.

\subsection{Discussion}
The following text discusses our insights learned from the experimentation and results. We also attempt to find out our limitations and subsequent future directions.

% \todo{Add a paragraph/subsection related to our findings on the edge-weight, why no improvement still now, how to improve using the information coming form edge weight? }
%Remember this part is not polished. Polish it carefully.
% \subsection{Is the Deep Model Always Good?
\par
\begin{figure}[h]
    \centering
    \includegraphics[width=0.65\textwidth,height=0.42\textwidth]{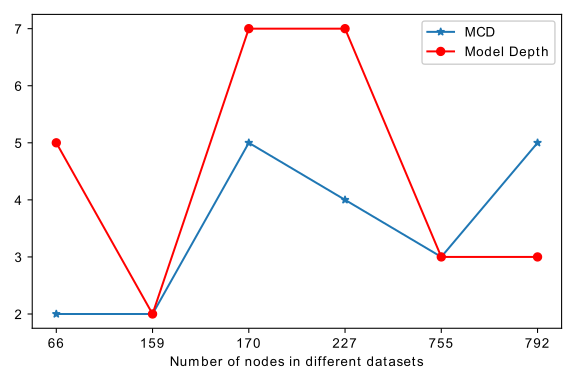}
    \caption{%\textcolor{red}{This figure shows 
    Maximum community diameter (MCD) with the best model depth. % for different datasets. Datasets are ordered from smaller to larger in terms of the number of nodes. It is clearly evident that MCD and model depth are very correlated. In fact, in some cases, model depth is exactly the same as MCD. 
    }
    \label{fig:mcdvsdepth}
\end{figure}
\textbf{Model Depth:} 
From the experiments on the different datasets, we find that the best model depth is subjective to the dataset. Sometimes, we find that the smaller datasets have the best %results
performance with a smaller number of layers, and comparatively larger datasets %such as the topics dataset
achieve the best results with a large number of layers. The reason behind this is that the smaller datasets often have smaller community %sizes/\textcolor{red}{
diameters while the larger datasets have larger community %sizes/\textcolor{red}{
diameters. It is evident that the deeper the GCN model, the more distant neighbors are aggregated. Therefore, the deep model can aggregate information from distant neighbors, which is required to discover communities %\textcolor{red}{
with a larger diameter.
Accordingly, when the community %size/\textcolor{red}{
diameter is smaller, we should explore closer neighbors which requires less deep models. %as shown in 
It is clearly evident %clearly shows
in Figure \ref{fig:mcdvsdepth} that the best model depth is highly correlated with the maximum community diameter (MCD). Even in some cases  model depth is exactly the same as MCD.   
The key insight here is that we should not define a fixed depth for the model beforehand, rather we should vary the depth of the model depending on the nature of the dataset. %This implies that the GCN model depth is subjective to the dataset. 
% \noindent

% \subsection{Is the Dynamic Dilated Aggregation Necessary?}

\textbf{Dynamic Dilated Aggregation:} 
However, then one question arises why should we need the small depth DynaResGCN model for the smaller datasets? In fact, the DynaResGCN model achieves superior performance than the NOCD model in the case of smaller datasets also. Here comes the power of the dynamic dilated aggregation scheme in the DynaResGCN model. Because of the dilated aggregation in the DynaResGCN model, it can aggregate information from the second hop neighbors in one layer of the GCN without increasing the effective number of neighbors (explained earlier). The dynamicity comes from the random sampling of the augmented neighbors, which allows exploring a number of new neighbors at every layer
as we perform a new sub-sampling of the neighbors at every layer. Therefore, the model gets some new information at every layer. In fact, the dilated aggregation mechanism also helps to alleviate the over-smoothing problem \cite{li2019deepgcns}. %Moreover, weighted dilated aggregation (in weighted graphs) allows to exploit the information underlying in the edge weights.
%Hence
Consequently, dynamic dilated aggregation is one of the key ideas to achieving better %results
performance in the DynaResGCN model. 

\textbf{Residual Connection:}
Although the dynamic dilated aggregation mechanism solves the over-smoothing problem and gives better learning capability, the deep GCN models would still fail to learn because of the vanishing gradient problem \cite{xu2018representation,li2019deepgcns}. The residual connection provides the skip connections in the computation path of the deep neural networks. This direct computation path allows a smooth backward flow of the gradients from the last layer to the beginning layers. Consequently, the gradients do not vanish when they reach the beginning layers. Therefore, the residual connection is essential to train deep GCN models which we use in our deep DynaResGCN model.

\textbf{Network Property:}
Another important thing is the property of the network being studied. For example, if the network is very dense (almost fully connected), then no method can find a community structure in the network because the network has essentially no community structure. This property of a network is called the information-theoretic detection threshold \cite{chen2017supervised-line-gnn,banks2016information-detection-threshold}, which is well defined in the stochastic block model of community detection \cite{abbe2017community-sbm}. In general, if the graph being studied is almost regular, then there will be no community structure at all; eventually, no algorithm can detect a community structure. Therefore, when we study community detection in some %graphs/
networks, we should also consider the network property of the detection threshold, though it is well defined only for the special stochastic block model \cite{abbe2017community-sbm}. 

\textbf{Evaluation Metric:}
The metric to determine the quality of the detected community is also crucial. In fact, using only a single metric is always highly risky because if the metric fails due to a corner case, there is no way to detect this failure. Therefore, having multiple metrics to detect different characteristics of the detected communities is of utmost importance. We alleviate this issue using multiple metrics that capture different characteristics of the detected communities. We refer to a method winning only when it performs better %/good
in terms of all the metrics considered.  

\textbf{Threshold Sensitivity:} 
We have already discussed that every dataset is different from every other dataset. Therefore, the cut-off value (threshold) of the community strength (community affiliation) to determine the community belonging (whether a node belongs to a community) should be varied based on the dataset. Using the same threshold for every dataset is not a clever idea. In our experiments, we vary the threshold in a specified range and determine the best threshold for every dataset. We can observe the best threshold for every dataset in Table \ref{tab:dyna-G-details} and Table \ref{tab:dyna-x-details}. In these tables, we can observe that the best threshold is not the same for all datasets. Therefore, it is necessary to consider variable thresholds in order to achieve the best result from every dataset. The key finding here is that the threshold value to determine the community belonging is subjective to the dataset being considered.

\textbf{Robustness:}
From the results in Table \ref{tab:ground-truth-G} and Table \ref{tab:ground-truth-x}, it is clearly evident that the DynaResGCN method is much more robust to node features than that of NOCD or other methods. In fact, the performance of DynaResGCN-G is much better than NOCD-G. We consider a one-hot encoded feature vector for every node when we do not have specific node features. As a result, the feature vector for each node is unique, which helps to achieve a deeper model without over-smoothing. When we have node features, the deeper model does not work well in the case of very large graphs. The most probable reason is the over-smoothing due to similar feature vectors of many nodes. Therefore, DynaResGCN achieves better performance without node features due to lower over-smoothing, and it also performs better with node features due to additional information coming from node features. These two advantages in the two cases make the DynaResGCN method much more robust than related methods.

\textbf{Statistical Significance:}
Our best method, DynaResGCN, is significantly better than the nearest best method in almost every case. We measure the significance based on a standard $t$-test \cite{t-test,t-test2} with $95\%$ confidence. In fact, very few works in the literature provided statistical significance for their results. Therefore, we can provably state that DynaResGCN is a significantly better method than many state-of-the-art methods for overlapping community detection.

\textbf{Scalability:}
GCN-based methods are extremely scalable for overlapping community detection. The execution time for DynaResGCN is less than 15 seconds, even with seven layers for small datasets. In this case, DynaResGCN does not impose any extra burden on the execution time. For datasets up to $22$ thousand nodes, DynaResGCN took less than 2 minutes of execution time with a deep model of up to 15 layers. In fact, the execution time of DynaResGCN is not more than 15 minutes, even with a very large dataset ($65K$ nodes) with the deep model. Therefore, our method, DynaResGCN, is highly scalable even with the very deep model. As a result, the GCN-based deep methods would be much more suitable than any other method in practice where real-time overlapping community detection is required and %even 
in the case where networks are very dynamic, such as social networks. 

\textbf{Limitation and Future Directions:}
First of all, it is crucial to incorporate the information from edge weights in the case of weighted graphs. Therefore, there are scopes to design methods and algorithms to incorporate information lying in the edge weights in a  meaningful and sophisticated way. In the very large datasets with node attributes, the DynaResGCN-based method achieves the best performance with only two layers in some cases. This implies that the DynaResGCN model can not perfectly resolve the over-smoothing problems. As a result, in some cases, deeper models do not outperform. Resolving the over-smoothing problem in better ways would be an interesting future research direction. In this study, we focus on improving the encoder part of the whole framework. There are scopes to research on improved decoders considering different properties of community. Finally, we apply our developed methods and algorithms in a completely unsupervised setting. It would be interesting to incorporate our ideas, methods, and algorithms in supervised settings of graph learning.
% In the case of  very large $(nodes>10K)$ and dense  graphs with feature vectors for the nodes, deep models are not outperforming the shallow model with large margin. Some cases, we get the best results in our DynaResGCN model with only two layers. Therefore, we can not state that the deeper models are always the best methods in case we have node features which may be co-related. 

% \textbf{Future Directions}
% Improvements on weighted graphs. Training deep GCN in very large graphs. Solving over-smoothing in a better way. Improving the decoder based on more community specific concepts. Apply these idea of deep GCNs in supervised settings.

% \subsection{What can we Improve through Residual Connection?}

% \subsection{Does the Graph Property Matter?}

\section{Conclusion}
In %our 
this study, we develop dynamic dilation aggregation-based deep residual graph convolutional network (DynaResGCN) to detect overlapping communities in graphs. We resolve %serveral 
several challenges to incorporate dilated aggregation, dynamic edges, %edge-weights, 
and residual connection in the general irregular graphs for overlapping community detection tasks. Eventually, our methods significantly outperform most of the prominent existing works. %However, it remains open to incorporate edge weights in the deep GCNs in the overlapping community detection setup. %The f
Future works should investigate the challenges of %heterogenous 
heterogeneous graphs where heterogeneity
%genety 
comes from both edges and nodes. %It is also worthy to investigate the 
The challenges of GNNs in %the 
very large networks, such as the whole Facebook or Twitter networks, are also worthy of investigation in the context of overlapping community detection. Moreover, consideration of edge-weights in a meaningful way can be an interesting future research direction.  %Our 
This work has %applications in 
many different applications, ranging from bioscience %bio-science 
to social science and %in 
any tasks where there is a network structure. 

\subsubsection*{Acknowledgements}
This work is supported under M. Sc. Engg. Fellowship provided by the ICT Division of Bangladesh
Government. For experimentation, we used the cloud computing facilities of the University of Arizona, USA.
\subsubsection*{Data Availability Statement}
The research topics dataset analyzed during the current study is available in the corresponding repository at \url{https://github.com/enggiqbal/mlgd/tree/master/data/datasets/topics/orginal}. Other analyzed datasets are available in the NOCD repository at \url{https://github.com/shchur/overlapping-community-detection/tree/master/data}, and our code is available at \url{https://github.com/buet-gd/Deep-DynaResGCN-community}.
\newpage
\bibliographystyle{splncs04}
\bibliography{main}

\end{document}